\documentclass[journal]{IEEEtran}
\usepackage{hyperref}
\usepackage{graphicx}
\usepackage{multirow}
\usepackage{amsmath,amssymb,amsfonts}
\usepackage{amsthm}
\usepackage{mathrsfs}
\usepackage{xcolor}
\usepackage{textcomp}
\usepackage{manyfoot}
\usepackage{booktabs}
\usepackage{algorithm}
\usepackage{algorithmicx}
\usepackage{algpseudocode}
\usepackage{listings,rotating}
\usepackage{subcaption}
\usepackage{graphicx}
\usepackage{tikz}
\usepackage{mdframed}

\ifCLASSINFOpdf
  
\else
  
\fi

\begin{document}

\title{Logic-Guided Multistage Inference for Explainable Multidefendant Judgment Prediction}

\author{Anonymous Authors
% \thanks{Manuscript submitted to IEEE Transactions on Neural Networks and Learning Systems.}
\thanks{All authors are affiliated with an anonymous institution. Email addresses and detailed affiliations are removed for double-blind review.}
\thanks{Funding information is omitted for anonymization and will be provided after the review process.}
}

\author{Xu Zhang, Qinghua Wang, Mengyang Zhao, Fang Wang, and Cunquan Qu

\thanks{\textit{Corresponding author}: Fang wang and Cunquan Qu}
\thanks{Xu Zhang, Qinghua Wang, Mengyang Zhao, Cunquan Qu, and Fang Wang are with the Data Science Institute, Shandong University, Jinan 250100, China (email: xzhang@sdu.edu.cn; qinghuawang@mail.sdu.edu.cn; mengyangzhao@mail.sdu.edu.cn; wangfang226@sdu.edu.cn; cqqu@sdu.edu.cn).}
\thanks{This work is supported in part by funds from the National Science Foundation (NSF: \#T2293773, \#T2293770 and \#12471488).}
}

\markboth{Journal of \LaTeX\ Class Files,~Vol.~X, No.~X, August~2025}%
{Shell \MakeLowercase{\textit{et al.}}: Bare Demo of IEEEtran.cls for IEEE Journals}

\maketitle

\begin{abstract}
Crime disrupts societal stability, making law essential for balance. In multidefendant cases, assigning responsibility is complex and challenges fairness, requiring precise role differentiation. However, judicial phrasing often obscures the roles of the defendants, hindering effective AI-driven analyses. To address this issue, we incorporate sentencing logic into a pretrained Transformer encoder framework to enhance the intelligent assistance in multidefendant cases while ensuring legal interpretability. Within this framework an oriented masking mechanism clarifies roles and a comparative data construction strategy improves the model's sensitivity to culpability distinctions between principals and accomplices. Predicted guilt labels are further incorporated into a regression model through broadcasting, consolidating crime descriptions and court views. Our proposed masked multistage inference (MMSI) framework, evaluated on the custom IMLJP dataset for intentional injury cases, achieves significant accuracy improvements, outperforming baselines in role-based culpability differentiation. This work offers a robust solution for enhancing intelligent judicial systems, with publicly code available\footnote{\url{https://github.com/XuZhang29/MMSI}}.
\end{abstract}


\IEEEpeerreviewmaketitle

\section{Introduction}
\IEEEPARstart{A} reasonable distribution of responsibilities is essential for maintaining a fair and orderly society \cite{rawls1958justice, corning2011fair, nisbet2023quest}. Individuals assume varied roles that, through coordinated efforts, contribute to the smooth functioning of society \cite{sunstein1996social, fafchamps2011development, nikulina2022interplay}. When the social order is disrupted, the establishment of agreed-upon rules is needed to restore it \cite{garland2001culture, duncan2010crime, henry2019crime}. In primitive societies, customary norms emerged as the earliest form of law \cite{hanba2022archaic}, evolving through harsh eras such as “tit for tat” \cite{roth2014eye}. Modern law prioritizes fairness and justice, emphasizing the integration of “law, reason, and emotion” into legislation and enforcement \cite{rawls2001justice, sellers2017law}. Thus, accurately identifying the culpability of individuals under the law and imposing reasonable and fair punishments are crucial for effective governance.

Guilt reasoning in judicial work requires judges to apply strict judicial logic and possess a comprehensive understanding of case facts \cite{pound1922theory}. However, the rapid increase in the number of judicial cases has created a shortage of professionals, emphasizing the urgent need for simple and efficient judicial assistance tools. While data-driven judicial tools such as COMPAS \cite{zhang2014analysis} and crime risk analytics have shown promise \cite{berk2016forecasting, kumar2020crime}, most rely on structured data and focus on narrow applications like recidivism prediction or fraud detection.
Recent studies further highlight practical constraints in real deployments, where key judicial findings may be unavailable at early stages, motivating predictions directly from raw case materials and incomplete evidence \cite{liu2025_lfp}.

Recent research has focused on modeling unstructured legal texts, including legal judgment prediction (LJP) \cite{cheng2020knowledge, gao2023legal, bi2023judicial}, legal question answering (LQA) \cite{zhong2020iteratively}, similar case matching (SCM) \cite{bi2022learning}, and legal text summarization (LTS) \cite{bhattacharya2019comparative}. The LJP task leverages judicial document data to enhance the predictions produced across various domains, including legal charge prediction (LCP) \cite{jiang2018interpretable, zhao2022charge, bi2024knowledge}, legal prison prediction (LPP) \cite{yue2021neurjudge, li2020prison, liu2023ml}, and legal article prediction (LAP) \cite{hu2018few, luo2017learning, xu2020distinguish}.
Alongside accuracy gains, recent efforts increasingly emphasize interpretability and robustness under distribution shifts (e.g., statute revisions and temporal drifts), which are critical for trustworthy judicial assistance \cite{li2025_iljr, han2025_lawshift}.

Multidefendant case prediction remains relatively underexplored compared with single-defendant settings, despite constituting a substantial portion of real-world cases \cite{pan2019charge}. Very recent work has started to explicitly address this gap using structured event representations and multi-agent coordination to model inter-defendant interactions \cite{yuan2026_magljp}. The interactions among defendants are complex, requiring clear distinctions among their roles, their actions, and the causal relationships between actions and violated rights. This complexity poses significant challenges for intelligent sentencing assistance methods, as it demands nuanced reasoning that is traditionally reliant on judges' deep insights.
Many existing models, such as LAP-based pipelines, fail to capture these individualized roles and often assume uniform legal articles, undermining prediction reliability. As legal roles significantly impact sentencing (e.g., principal: +0.0387, accomplice: –0.344 \cite{wang2022applications}), they must be explicitly modeled \cite{hu2020identifying, lyu2023multi}.
According to Articles 26 and 27 of the Criminal Law of the People's Republic of China\footnote{\url{http://xingfa.org/}}, principals are defined as the main actors in joint crimes, whereas accomplices play minor or supportive roles. These roles are not merely descriptive but fundamentally affect judicial outcomes and thus must be computationally modeled with care.

To address these challenges, we infer principal and accomplice guilt labels from criminal facts with a defendant-oriented masking mechanism that sharpens role-specific representations. We then broadcast these inferred labels to the sentencing stage and embed them into a prediction model that follows the judicial process of first determining guilt and then deciding the sentence. This design enhances role identification and interpretability in multidefendant cases and improves sentencing accuracy while remaining compatible with the fixed input length of the underlying pretrained Transformer encoder.
The experimental results demonstrate the efficacy of the proposed method in assisting guilt differentiation and providing sentencing recommendations for multidefendant cases. Moreover, it outperforms state-of-the-art (SOTA) large language models (LLMs) in this domain.

Building on this setting, the main contributions of this work are summarized as follows:
(1) A masked multistage inference framework aligning guilt determination and sentencing.
(2) Defendant-oriented masking and comparative data construction for robust role differentiation.
(3) A label broadcasting mechanism that enhances sentencing prediction beyond multitask learning.
(4) The IMLJP multidefendant dataset and comprehensive evaluations, including LLM baselines and interpretability analyses.

The remainder of this work is structured as follows. Section~\ref{sec:related_work} reviews related work. Section~\ref{sec:method} introduces the proposed methods, including oriented masking, task-specific data usage, and a multitask cascading framework for sentencing. Section~\ref{sec:experiments} presents dataset construction and model evaluations, accompanied by interpretability analysis. Section~\ref{sec:conclusions} concludes the study and outlines future directions. Additional results, including ablation and sensitivity analyses, are provided in the Appendix.

\section{Related Work}
\label{sec:related_work}
Early research in LJP mainly focused on single-defendant cases, typically predicting charges, applicable law articles, and prison terms.
TopJudge \cite{zhong2018legal} introduced a multitask learning framework for synchronous predictions across these subtasks. NeurJudge \cite{yue2021neurjudge} further improved predictions by leveraging intermediate outputs from subtasks to refine final judgments. LADAN \cite{xu2020distinguish} integrates a graph distillation operator to disentangle confusing statutes, improving both prediction accuracy and interpretability. ML-LJP \cite{zhang2023contrastive} employs contrastive learning and Graph Attention Networks (GAT) to capture inter-law relationships and better handle numeric evidence in sentencing prediction.
K-LJP \cite{li2025legal} incorporates both label-level knowledge (relations between legal labels) and task-level knowledge (alignment across charges and law articles), enabling richer synergies across tasks.
Beyond multi-task architectures, recent work has explicitly targeted interpretability by grounding judgment reasoning in multi-source knowledge and producing explanation-friendly rationales \cite{li2025_iljr}.
Additionally, pretrained legal language models such as Legal-BERT \cite{chalkidis2020legal} and Lawformer \cite{xiao2021lawformer} provide strong backbone representations, offering domain-specific understanding and long-text modeling capabilities, though they were not originally designed specifically for LJP.
Robustness has also become a growing concern, with recent benchmarks evaluating LJP under statute shifts to quantify performance degradation in realistic evolving legal environments \cite{han2025_lawshift}.

Multidefendant cases involve overlapping actions, intertwined causal relationships, and distinct legal roles, which pose additional challenges. Early work attempted to address this through shallow feature modeling. Hu et al. \cite{hu2020identifying} used behavioral semantics and statistical features to model defendant relationships. However, their reliance on sentence complexity features limits the ability of the model to conduct deeper semantic analyses. Andreas et al. \cite{andreas2023counterfactuals} proposed counterfactual disturbances to clarify defendant roles, but this approach is limited by its reliance on idealized social structures. Lyu et al. \cite{lyu2023multi} employed hierarchical reasoning framework (HRN) for classifying actions in multidefendant cases, a two-level framework that first identifies inter-defendant relationships and then predicts each defendant’s charges, law articles, and prison terms, using Seq2Seq with mT5 and Fusion-in-Decoder for long case descriptions.
More recently, structured event representations and multi-agent coordination have been explored to explicitly model inter-defendant interactions for multi-defendant LJP \cite{yuan2026_magljp}.

LLM-based adaptations have become a new frontier in LJP, where prompting and fine-tuning are often combined.
Huang et al. \cite{huang2024cmdl} introduced the CMDL dataset for multi-defendant scenarios and evaluated prompting methods with DependantT5, a T5-based model exploiting subtask dependencies.
Vanilla-SFT unifies training data into chat templates, while Finetune-CoT \cite{ho2024large} leverages synthesized reasoning chains to strengthen inference.
Other approaches such as CL4LJP \cite{zhang2023contrastive} and CECP \cite{zhao2022charge} show promise but remain less adaptable to multi-label classification.
ADAPT \cite{xu2024distinguish} structures LJP tasks into multi-stage reasoning trajectories, effectively fine-tuning models using synthetic chains of inference. GLARE \cite{yang2025glare} implements agentic reasoning to dynamically invoke legal knowledge modules during inference, balancing interpretability and adaptability.
Complementary directions incorporate symbolic constraints or logical rules into LLM-based prediction to better align outputs with legal reasoning patterns \cite{zhang2025_rljp, wei2025_neurosymbolic}, while practical pipelines increasingly consider missing or delayed judicial findings by introducing intermediate targets such as legal fact prediction \cite{liu2025_lfp}.
These approaches demonstrate that LLMs and prompt strategies, although not legal-specific models, can effectively scale LJP systems and generalize to complex multi-defendant scenarios.

\section{Methodology}
\label{sec:method}
\begin{figure*}[!htbp]
	\centering
	\includegraphics[width=0.85\linewidth]{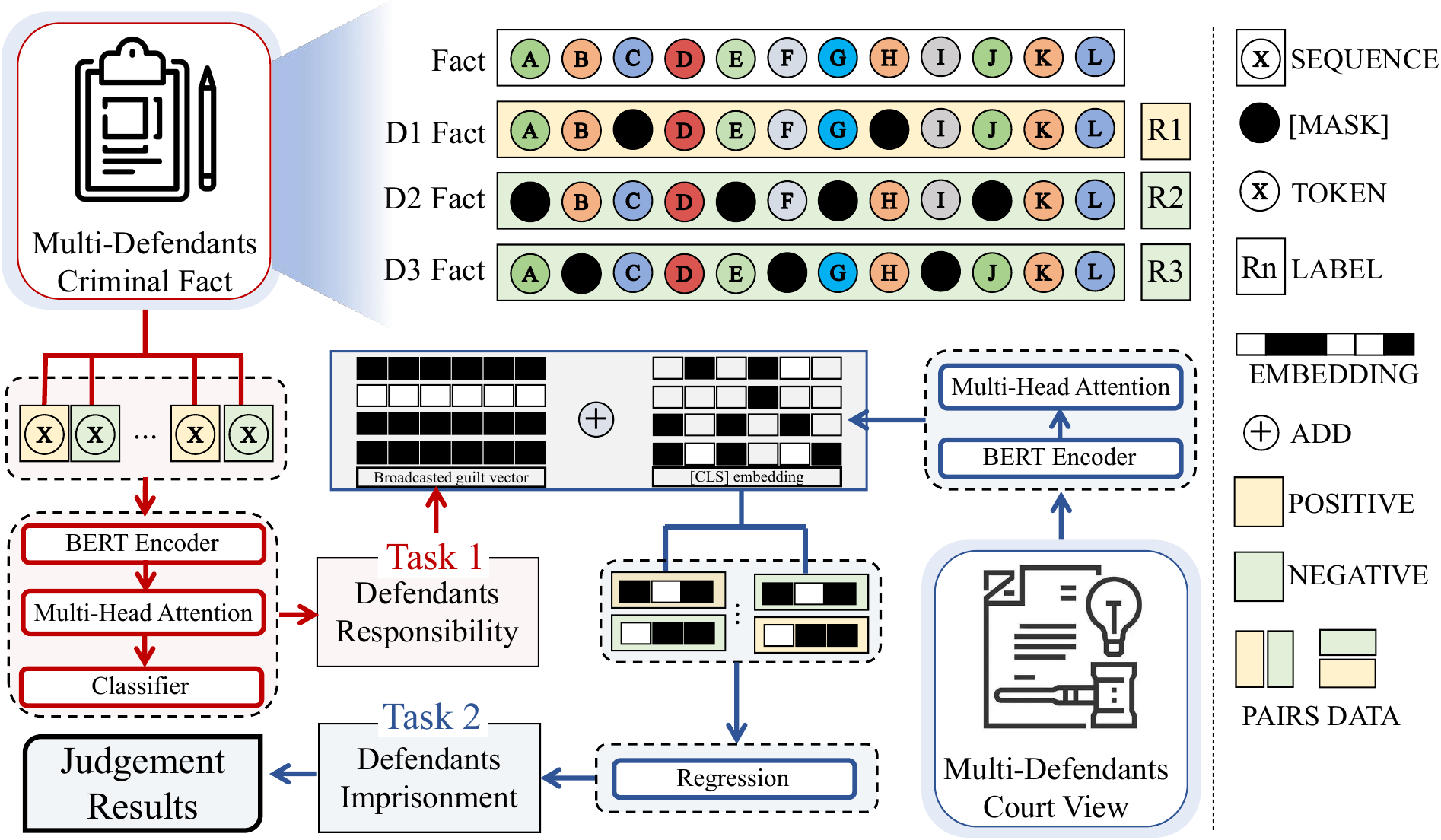}
	\caption{\label{fig:framework}
    Framework of the proposed MMSI for multidefendant cases. This approach involves two primary tasks: Task 1 (shown in red), which performs defendant guilt--responsibility reasoning on the basis of fact descriptions (FDs), and Task 2 (shown in blue), which predicts sentencing terms via both FDs and pruned court views (CV$_d$, i.e., court views with guilt-role-related sentences removed). Both the FD and CV$_d$ inputs undergo preprocessing with an oriented masking technique, and the training data are enhanced through contrastive data construction, which is applied to both tasks to achieve improved prediction accuracy. In Task 2, the guilt responsibility labels derived from Task 1 are broadcast within the model and embedded into text representations, facilitating the integration of multisource data into MMSI for prison prediction.}
\end{figure*}

The proposed masked multistage inference (MMSI) framework is illustrated in Fig.~\ref{fig:framework}. Specifically, text data, including fact descriptions (FDs) and their pruned court views (CV$_d$), undergo preprocessing with oriented masking and contrastive dataset construction techniques. The encoding process employs a bidirectional Transformer encoder with a multihead attention mechanism to support both guilt inference and imprisonment prediction tasks. 
Notably, this approach treats guilt inference as an auxiliary stage for prison prediction. Instead of concatenating FDs and CV$_d$, it propagates the inferred guilt labels from FDs to CV$_d$ via label broadcasting, which alleviates the sequence length limitations of Transformer-based encoders on long judicial documents.

%Notably, this approach treats the guilt inference task as a subtask within the prison prediction task, integrating FDs into CV$_d$s through label propagation techniques, which helps address the sequence length limitations of Transformer-based encoders when processing long judicial documents.

When converting text data into a computer-readable format, encoding plays a crucial role. The emerging pretrained language models \cite{vaswani2017attention, howard2018universal}, particularly base models that are fine-tuned on extensive corpora, can effectively preserve the semantic information of the original text, streamlining this process. However, the performance of these pretrained models in semantic encoding tasks relies heavily on the richness and professionalism of the utilized corpus. 
Using the guilt inference task as a representative example, we compare several widely used pretrained models for legal text tokenization. Among them, the BERT-base-Chinese \cite{jacob2018DBLPchinese} tokenizer and embedding layer achieve the best performance on the guilt inference task defined in Section~\ref{sec:problem_formulation}, so they are adopted as a unified front end for all models (see Table~\ref{tab:model_compare_tasks} in Appendix~\ref{appsec:A_tokenizer}).

Note that, given the inherent contrast between principals and accomplices in multidefendant cases, all training datasets for principal and accomplice classification experiments are constructed by using comparative data pairs. Details of the comparative data construction procedure and the corresponding performance validation are provided in Appendix~\ref{appsec:B_consdata}.

\subsection{Logic and Data for LJP Tasks}\label{subsubsec2}
\begin{figure}[!ht]
	\centering
	\includegraphics[width=0.8\linewidth]{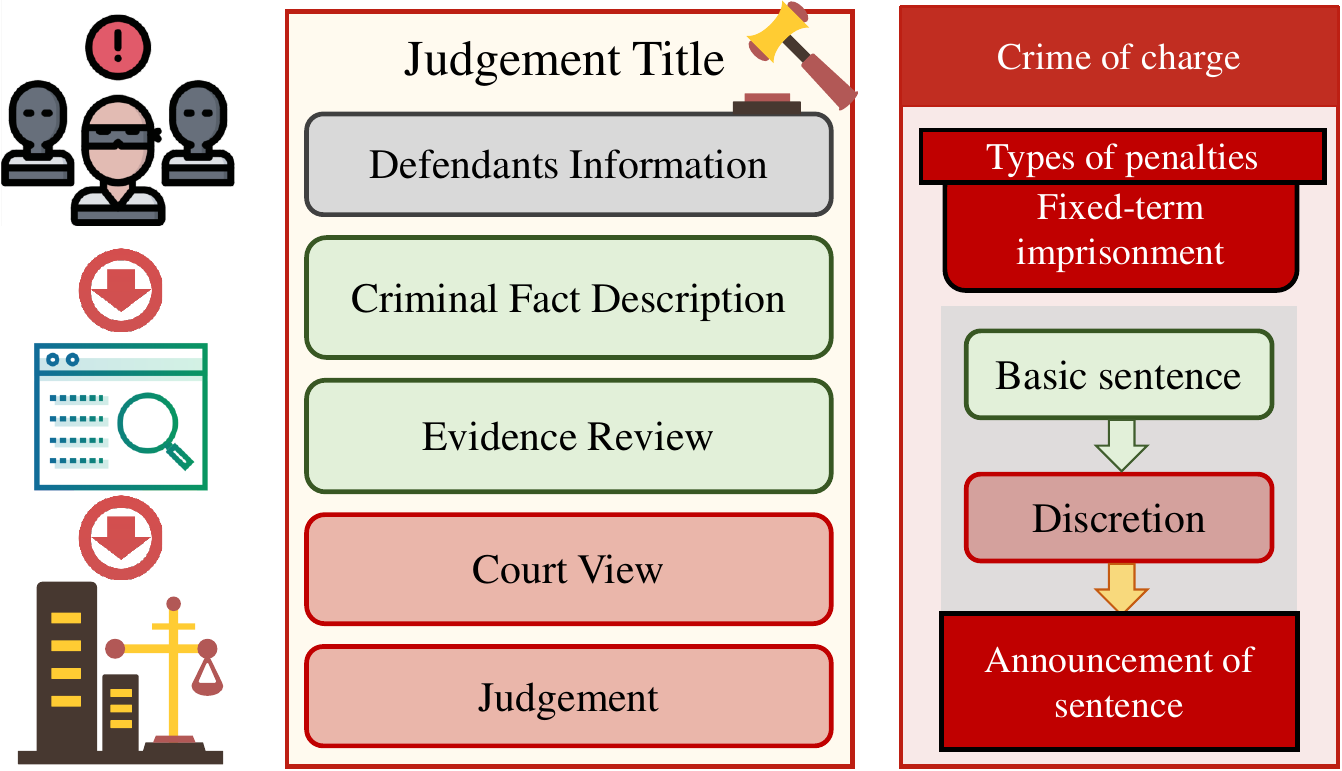}
	\caption{\label{fig:judgement_process}Structures of judicial documents and sentencing logic. The left panel illustrates the temporal sequence consisting of the crime, trial, and sentencing stages. The middle panel presents the composition of a judicial document. The right panel outlines the sentencing logic process for determining prison terms under a specific criminal charge.}
\end{figure}

Most existing judgment prediction models lack transparency regarding the judge’s decision-making process—an interpretability gap this work seeks to address. As illustrated in the left panel of Fig. \ref{fig:judgement_process}, the judicial process begins with a criminal act committed by a group, followed by law enforcement investigations and arrests. The procuratorate initiates a case based on collected evidence and witness testimonies, and the court ultimately issues a public judgment. 

The structure of a typical judgment document, illustrated in the middle panel of Fig.~\ref{fig:judgement_process}, consists of five sections: defendant information (DI), fact description (FD), evidence review (ER), court view (CV), and the final judgment. The first three sections are drafted by the procuratorate, while the CV and judgment are composed by the judge. The CV synthesizes the established criminal FD, discusses mitigating or aggravating factors (e.g., confession, surrender), and applies the relevant legal provisions. The final judgment includes charges, types and durations of penalties, execution methods (e.g., probation), and fines. Among various punishment types, fixed-term imprisonment is of particular interest due to its quantifiable nature and social impact, and thus serves as the focus of this study. 

The right panel of Fig.~\ref{fig:judgement_process} depicts a two-stage sentencing process: (1) determining a base term from applicable legal articles and case specifics, and (2) adjusting the sentence using discretionary factors such as voluntary surrender, confession, recidivism, and the defendant’s role (principal vs. accomplice).

According to the Guidelines on Sentencing for Common Crimes\footnote{\url{http://gongbao.court.gov.cn/Details/0b0f4c58cf4a9d00abdf5ffa0b3eb5.html}}, judges must consider repentance, surrender, and other contextual factors documented in the CV when determining the final sentence. Therefore, regression based on CVs is essential for fixed-term sentencing prediction. However, prior approaches relying solely on FDs often fail to capture these nuanced adjustments. 

In multidefendant scenarios, the FD often presents collective behaviors without clarifying individual responsibilities or causal links. For instance, in a group assault case, if one defendant inflicts the fatal injury, the FD typically emphasizes that individual’s role. Judges must interpret such interactions to assign principal or accomplice labels—a process requiring substantial legal reasoning. To support this process, AI models must infer roles directly from the FD. Although CVs may contain such annotations, their varied linguistic expressions hinder reliable rule-based extraction. This work addresses both guilt inference from FDs and guilt identification from CVs, aiming to reduce dependence on limited judicial resources.

\subsection{Problem Formulation}
\label{sec:problem_formulation}
We consider three prediction tasks in multidefendant cases: 
(i) guilt inference from FDs, 
(ii) guilt identification from CVs, and 
(iii) prison prediction when principal/accomplice–related sentences have been removed from the CVs. 
This setup mirrors judicial practice, where judges infer roles from the FD, record them in the CV, and then determine prison terms. 
It also allows us to analyze three aspects separately: how well roles can be inferred directly from FDs (guilt inference), how effectively different roles can be distinguished when explicit role information is present in CVs (guilt identification), and how sentencing behaves when explicit role information is unavailable (prison-term prediction). In the MMSI framework, only guilt inference and prison-term prediction are jointly modeled. The guilt-identification task is used solely in standalone experiments to assess the model’s ability to distinguish principal and accomplice roles from CVs with explicit role cues and is not part of the MMSI architecture.

Each judicial document contains FD and CV, which are represented as token sequences divided by characters:  
\[
\text{FD} = \{t_1, t_2, \dots, t_m\}, \quad \text{CV} = \{t'_1, t'_2, \dots, t'_n\},
\]
where $m$ and $n$ denote the lengths of the respective sequences. Each case also includes a set of defendant names $D = \{d_1, d_2, \dots, d_l\}$, with $l$ indicating the number of defendants. In this formulation, each $d_i$ is treated as a single atomic token representing the $i$-th defendant’s name, regardless of its surface length.

To simulate scenarios in which explicit judicial reasoning about guilt roles is unavailable, all sentences in the CV that mention principal or accomplice are removed, denoted as $\text{CV}_{d}$. This allows the sentencing model to be evaluated under conditions where guilt annotations are missing.

Formally, let \(y_g\), \(y_i\), and \(y_p\) denote the guilt inference, guilt identification, and sentencing labels, respectively. For each defendant \(d_i \in D\), the corresponding prediction functions are:
\[
\hat{y}_g = \phi_g(\text{FD}, d_i), \quad
\hat{y}_i = \phi_i(\text{CV}, d_i), \quad
\hat{y}_p = \phi_p(\text{FD}, \text{CV}_d, d_i),
\]
where \(\phi_g\), \(\phi_i\), and \(\phi_p\) predict the respective labels for defendant \(d_i\) based on FD, CV, and FD with \(\text{CV}_d\).

\subsection{Oriented Masking Method for Role-Specific Text Representations}
In complex multidefendant case descriptions, the interconnections among defendants often make it challenging to distinguish them via simple sentence segmentation strategies. The existing approaches address this challenge through methods such as sentence complexity analysis \cite{hu2020identifying} and generative models \cite{lyu2023multi}. 

\begin{figure}[!ht]
	\centering
	\includegraphics[width=0.95\linewidth]{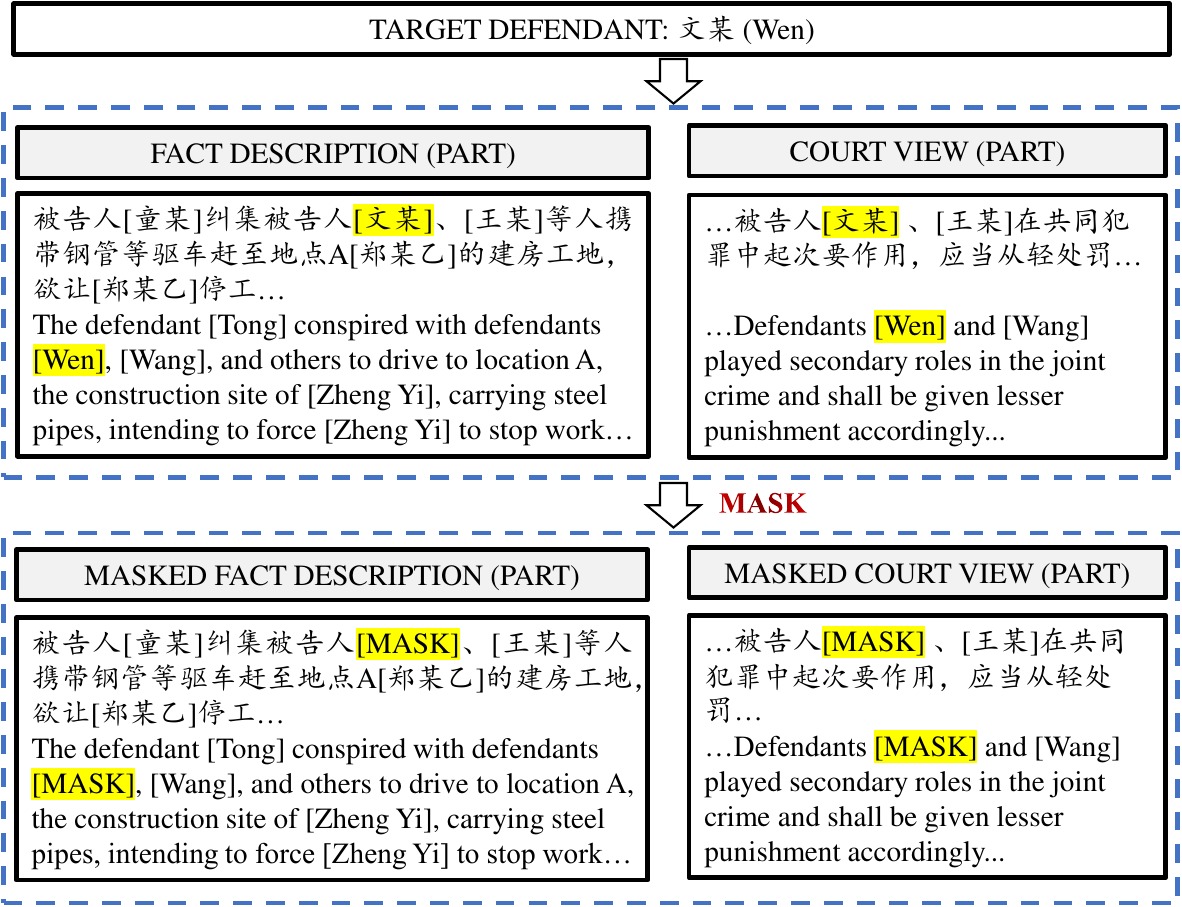}
	\caption{\label{fig:mask_example}Example of the oriented masking strategy. Based on the target defendant contained in the input, the left panel illustrates the masking process applied to the FD, whereas the right panel applies masking to the CV for the same defendant.}
\end{figure}

This work introduces a masking technique that differentiates textual descriptions of individual defendants and enhances their respective embeddings (see Fig. \ref{fig:mask_example} for an example). Specifically, a masking operation is applied to the token sequence of the defendant being analyzed, replacing the target defendant's name with a special token: [MASK]. The encoder thus learns role-specific representations conditioned on a unified role anchor. Unlike standard masked language modeling, [MASK] is not predicted but used as a structural marker that normalizes different name strings and emphasizes contextual cues of culpability.

To evaluate the effect of masking strategies in isolating defendant-specific features, we conduct experiments using a simple BERT-based classification model. This setup treats guilt inference and identification as independent classification tasks based solely on the provided input sequences. This setup follows the first-stage architecture in Section~\ref{sec:architecture}. This work designs three data preprocessing strategies for comparative analysis:

\begin{enumerate}
\item \textbf{Original}: The original input retains all defendant names in the FDs and CVs.
\item \textbf{Split}: Sentences are filtered to include only those involving the target defendant names.
\item \textbf{MASK}: All occurrences of the target defendant's name in FDs and CVs are replaced with the [MASK] token.
\end{enumerate}

These preprocessing techniques are evaluated on both the guilt inference task and the guilt identification task to assess how masking influences the model’s ability to distinguish principal and accomplice roles. The corresponding evaluation metrics are defined in Section~\ref{sec:experimental_settings}. Results of the guilt inference task are presented in Table~\ref{tab:MASK_effect}, with further results and analysis of the guilt identification task provided in Appendix~\ref{appsec:C_extr_FD_exp}.

\begin{table}[!htbp]
	\centering
	\caption{Oriented MASK effectiveness comparison in guilt inference tasks.}
	\label{tab:MASK_effect}
	\begin{tabular}{l|cccc}
		\hline
		\multirow{2}{*}{Method} & Acc. & P. & R. & F1. \\
		& ($\pm$95\%CI) & ($\pm$95\%CI) & ($\pm$95\%CI) & ($\pm$95\%CI) \\
		\hline
		\multirow{2}{*}{Original} & 0.5000 & 0.4126 & 0.3871 & 0.3539 \\
		& ($\pm$0.0000) & ($\pm$0.1284) & ($\pm$0.0062) & ($\pm$0.0872) \\
		\hline
		\multirow{2}{*}{Split} & 0.6281 & 0.6490 & 0.6238 & 0.6351 \\
		& ($\pm$0.0039) & ($\pm$0.0149) & ($\pm$0.0052) & ($\pm$0.0062) \\
		\hline
		\multirow{2}{*}{MASK} & \textbf{0.9105} & \textbf{0.9243} & \textbf{0.8998} & \textbf{0.9117} \\
		& ($\pm$0.0029) & ($\pm$0.0056) & ($\pm$0.0052) & ($\pm$0.0028) \\
        \hline
        \multicolumn{5}{r}{\footnotesize{$\pm$95\% CI: 95\% confidence interval}}\\
	\end{tabular}
\end{table}

The results demonstrate that the proposed oriented masking mechanism substantially improves the identification and inference accuracy over that attained with unprocessed data and data processed through the conventional sentence segmentation strategy. This improvement suggests that applying the oriented masking method to defendant names effectively enhances the ability of the model to discern the role relationships among multiple defendants.

\subsection{Key Components of Judgment Prediction}
Sentencing is a fundamental aspect of judicial judgment with direct implications for personal freedom. In multidefendant cases, however, overlapping actions among defendants make sentencing particularly challenging. Given its continuous nature, sentencing is better suited to a regression approach than a classification approach, as treating sentencing as a classification task \cite{yue2021neurjudge, xu2020distinguish} often leads to an excessive number of labels and larger error margins. While traditional analyses emphasize the FD of a crime, additional factors reflected in the CV, such as a defendant's confession and prior offenses, play essential roles in sentencing decisions and merit thorough consideration.

% To substantiate these findings, hypothesis testing is conducted under the initial assumption of no correlation between guilt and sentencing (see Table \ref{tab:correlation}). However, both the Pearson and Spearman tests reveal a significant association. Additionally, FDs and CVs are compared for the prison regression task to determine the optimal data source. On the basis of these insights, a label-enhanced prediction model incorporating guilt labels as key features is developed, and its effectiveness in prison prediction tasks is assessed.

% \begin{table}[!htbp]
% 	\centering
% 	\caption{\label{tab:correlation}Hypothesis testing results concerning the correlation between guilt labels an sentencing.}
%     \setlength{\tabcolsep}{3pt}
% 	\begin{tabular}{lcc}
% 		\toprule
% 		\textbf{Correlation Type} & \textbf{Correlation Coefficient} & \textbf{P-value} \\
% 		\midrule
% 		Pearson & 0.0425 & $2.1590 \times 10^{-15}$ \\
% 		\addlinespace
% 		Significance &  &  \\
% 		\quad (Pearson) & Reject null hypothesis & Significant correlation exists \\
% 		\midrule
% 		Spearman Rank & -0.0359 & $2.0198 \times 10^{-11}$ \\
% 		\addlinespace
% 		Significance &  &  \\
% 		\quad (Spearman) & Reject null hypothesis & Significant correlation exists \\
% 		\bottomrule
% 	\end{tabular}
% \end{table}

To evaluate how different input sources contribute to sentencing prediction, we use a regression model that takes either the FD or the filtered CV as input. Note that this setup focuses solely on assessing input effectiveness; the full regression branch architecture is detailed in the next section. The evaluation metrics are defined in Section~ \ref{sec:experiments}. We compare three model variants, each utilizing different input sources:
\begin{enumerate}
	\item Prediction based on FD: This baseline model uses an FD as its input, with prison terms employed as labels for model training.
	\item Prediction based on CV: This baseline model uses a CV as its input, with prison terms utilized as labels for model training.
	\item Guilt label-enhanced CV (CV+Feat): This enhanced model incorporates guilt label broadcasting, using a CV as its input and prison terms as labels for model training.
\end{enumerate}
The corresponding results are summarized in Table~\ref{tab:prison_predict_component}.
\begin{table}[!htbp]
	\centering
	\caption{Exploration of key input components and influential factors in prison prediction.}
	\label{tab:prison_predict_component}
	\begin{tabular}{c|ccc}
		\hline
		\multirow{2}{*}{Data for Training} & \multicolumn{1}{c}{ImpScore} & \multicolumn{1}{c}{ImpAcc} & \multicolumn{1}{c}{ImpErr}\\
		& ($\pm$95\%CI) & ($\pm$95\%CI) & ($\pm$95\%CI) \\
		\hline
		\multirow{2}{*}{FD} & 0.7048 & 0.4054 & 0.0721\\
		& ($\pm$0.0873) & ($\pm$0.1395) & ($\pm$0.0307)\\
		\hline
		\multirow{2}{*}{CV} & 0.7523 & 0.4752 & 0.0567\\
		& ($\pm$0.0815) & ($\pm$0.1419) & ($\pm$0.0234)\\
		\hline
		\multirow{2}{*}{CV+Feat} & 0.7705 & 0.4851 & 0.0519\\
		& ($\pm$0.0765) & ($\pm$0.1420) & ($\pm$0.0223)\\
		\hline
        \multicolumn{4}{r}{\footnotesize{$\pm$95\% CI: 95\% confidence interval}}\\
	\end{tabular}
\end{table}

The comparative results imply that using CVs improves the accuracy of prison prediction, as they encapsulate the key elements that are necessary for this task. This suggests that traditional reliance on FDs alone is insufficient; a comprehensive approach must include the interpretation of the target case by the court and the guilt and responsibility of each defendant. The proposed label broadcasting method effectively utilizes these critical sentencing factors, thereby enhancing the resulting prison prediction accuracy.

\subsection{Multitask Integration}
\label{sec:architecture}
\begin{figure}[!ht]
	\centering
	\includegraphics[width=0.95\linewidth]{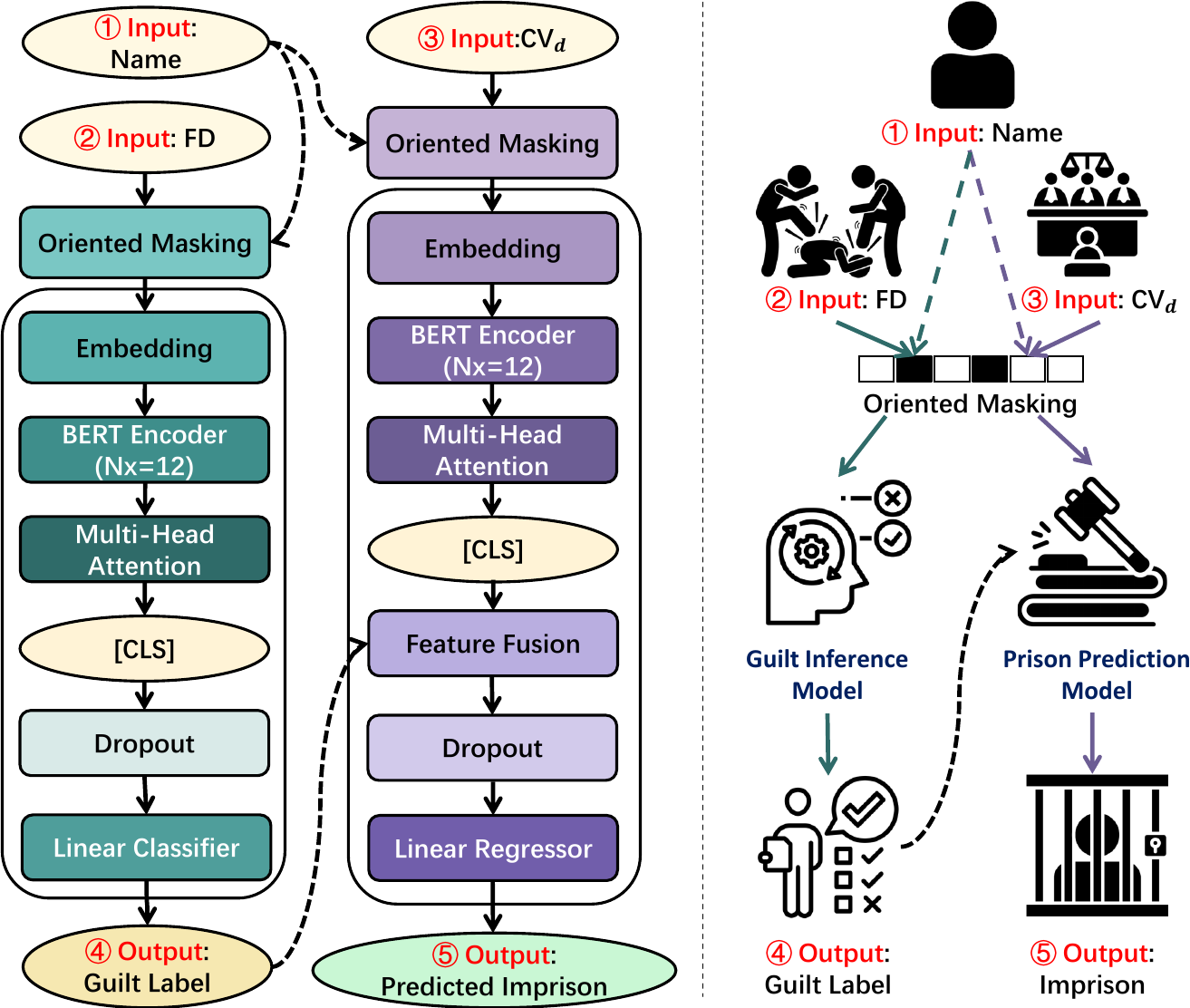}
	\caption{\label{fig:end2end_diagram}The architecture of the proposed MMSI framework and its correspondence to real-world judicial procedures. The left panel illustrates the end-to-end structure of MMSI, which jointly performs guilt inference and prison prediction based on three textual inputs: Name, FD, and CV$_{d}$. Through oriented masking, the model isolates the target defendant in multidefendant cases and infers guilt from FD using a classification branch. The inferred guilt label is then integrated into a regression branch that combines CV$_{d}$ and Name to predict the imprisonment term. The Embedding modules refer to the word-level embeddings produced by BERT tokenizer and its input embedding layer. The right panel depicts the real-world judicial reasoning process that MMSI emulates, including guilt inference from FD and sentencing based on CV$_{d}$. The oriented masking mechanism serves as a computational proxy for the judge’s focus on the target defendant.}
\end{figure}

The previously discussed model enhancement implemented through verified guilt labels significantly improves the accuracy of the model’s predictions. The final task of this work focuses on a cascaded auxiliary prison prediction approach for determining the guilt levels and sentences of multiple defendants. The overall architecture, including the broadcast-based fusion mechanism and the two-branch design, is illustrated in Fig.~\ref{fig:end2end_diagram}.

This framework follows a two-stage design:

\paragraph{Stage 1. Guilt Inference}
MMSI begins by predicting the guilt label for the target defendant using the FD and a masked representation of the defendant name. A 12-layer BERT encoder processes the input sequence, and a classification head predicts the binary guilt label \( \hat{y}_g \in \{0, 1\} \), indicating whether the defendant is a principal (1) or an accomplice (0). The model is trained by minimizing binary cross-entropy loss:
\[
\mathcal{L}_{\text{BCE}} = - \frac{1}{N} \sum_{i=1}^N \left[ y_{g,i} \log(\hat{p}_i) + (1 - y_{g,i}) \log(1 - \hat{p}_i) \right],
\]
where \( \hat{p}_i \) is the predicted probability for the \(i\)-th sample and \( y_{g,i} \) is the true label.

\paragraph{Stage 2. Sentencing Regression}
The predicted guilt label \( \hat{y}_g \in \{0,1\} \) is broadcast to match the dimensionality of the contextual representation derived from the CV$_{d}$. The CV$_{d}$ is processed using a 12-layer BERT encoder with Name-guided oriented masking, which emphasizes information relevant to the target defendant. To integrate guilt-related information into the sentencing model, the scalar prediction \( \hat{y}_g \) is expanded into a vector \( \mathbf{v}_g = \mathbf{1} \cdot \hat{y}_g \in \mathbb{R}^d \), where \( \mathbf{1} \) is a vector of ones with the same dimension \( d \) as the [CLS] embedding (i.e., the final hidden representation of the [CLS] token summarizing the CV$_d$ input). This broadcasted guilt vector is then added element-wise to the [CLS] embedding of the CV$_{d}$ representation:
\[
h_{\text{fused}} = h_{\text{[CLS]}}^{\text{CV}_d} + \mathbf{v}_g.
\]

The resulting fused embedding \( h_{\text{fused}} \) encodes both the sentencing context and the inferred guilt information of the defendant. A regression head then maps this representation to a predicted prison term \( \hat{y}_p \in \mathbb{R} \), which is optimized by minimizing the Mean Squared Error (MSE) between the predictions and ground-truth prison terms:
\[
\mathcal{L}_{\text{MSE}} = \frac{1}{N} \sum_{i=1}^N \left( y_{p,i} - \hat{y}_i \right)^2.
\]

In summary, MMSI first predicts the defendant's guilt label from the FDs and then injects this label into the sentencing regressor to obtain the final prison term, alleviating input-length constraints and integrating key information from multiple sections of the judgment document in line with judicial sentencing logic.

\section{Experiments}
\label{sec:experiments}

\subsection{Dataset Construction}\label{subsec2}

The Chinese Legal AI Challenge\footnote{\url{https://laic.cjbdi.com/}} (LAIC), launched in 2018, has significantly advanced LJP by using officially published datasets \cite{xiao2018cail} containing information on articles, charges, and sentences. However, these datasets predominantly cover single-defendant cases, resulting in limited resource availability for multidefendant scenarios. Lyu et al.'s dataset \cite{lyu2023multi} includes mixed crime data but does not clearly separate FDs from other content, introducing noise. While Huang et al. \cite{huang2024cmdl} proposed a large-scale multidefendant dataset (CMDL) with well-structured modular content, it does not annotate the relative guilt responsibility (e.g., principal vs. accomplice) of each defendant—a key focus of this work. As a result, CMDL is not directly applicable to our task, though we report benchmark results on it using our classifier model in Appendix~\ref{appsec:D_cmdl}. These gaps underscore the need for datasets that are specifically designed to address the complexities of multidefendant scenarios. From a sentencing standpoint, different crimes involve unique sentencing factors, necessitating clean datasets that are tailored to single-charge prediction tasks. Consequently, the existing datasets fall short in this regard. To bridge this gap, we focused on cases involving the crime of intentional injury and restricted our work to first-instance judgments. We constructed a multidefendant LJP dataset, referred to as \textbf{IMLJP}, which covers cases ranging from 2012--2020 and sourced from publicly available legal documents on the China Judgment Online website\footnote{\url{https://wenshu.court.gov.cn/}}. The dataset is available as part of our code repository.

In IMLJP, the keywords “principal” and “accomplice” were ensured to be included in the original CVs. Utilizing regular expressions, we extracted the FD, CV, and final judgment details for each case, as well as the defendants' names, sentences, probation statuses, and probation durations. To maintain data security, Ernie-Bot 3.5 within China was used to annotate the principal and accomplice labels within the CVs, with all personal identifiers (such as names and locations) anonymized. The dataset was subsequently reviewed and corrected by three professional annotators. After conducting preprocessing and manual annotation, IMLJP included 17,253 cases and 34,828 defendants, each annotated with verified principal and accomplice labels. A summary of the keyword information contained in IMLJP appears in Table \ref{tab:dataset_information}.
\begin{table}[!htbp]
	\centering
	\caption{Dataset information.}
	\label{tab:dataset_information}
	\begin{tabular}{l|l}
		\hline
		\textbf{Key} & \textbf{Description} \\
		\hline
		@id & Unique case ID \\
		@name & Defendant name \\
		@FD & Criminal fact description \\
		@CV & Court view \\
		@prison & Prison sentence (months) \\
		@probation & Probation duration (months) \\
		@guilt & Defendant role (principal/accomplice) \\
		\hline
	\end{tabular}
\end{table}

\begin{figure}[!ht]
	\centering
	\includegraphics[width=1\linewidth]{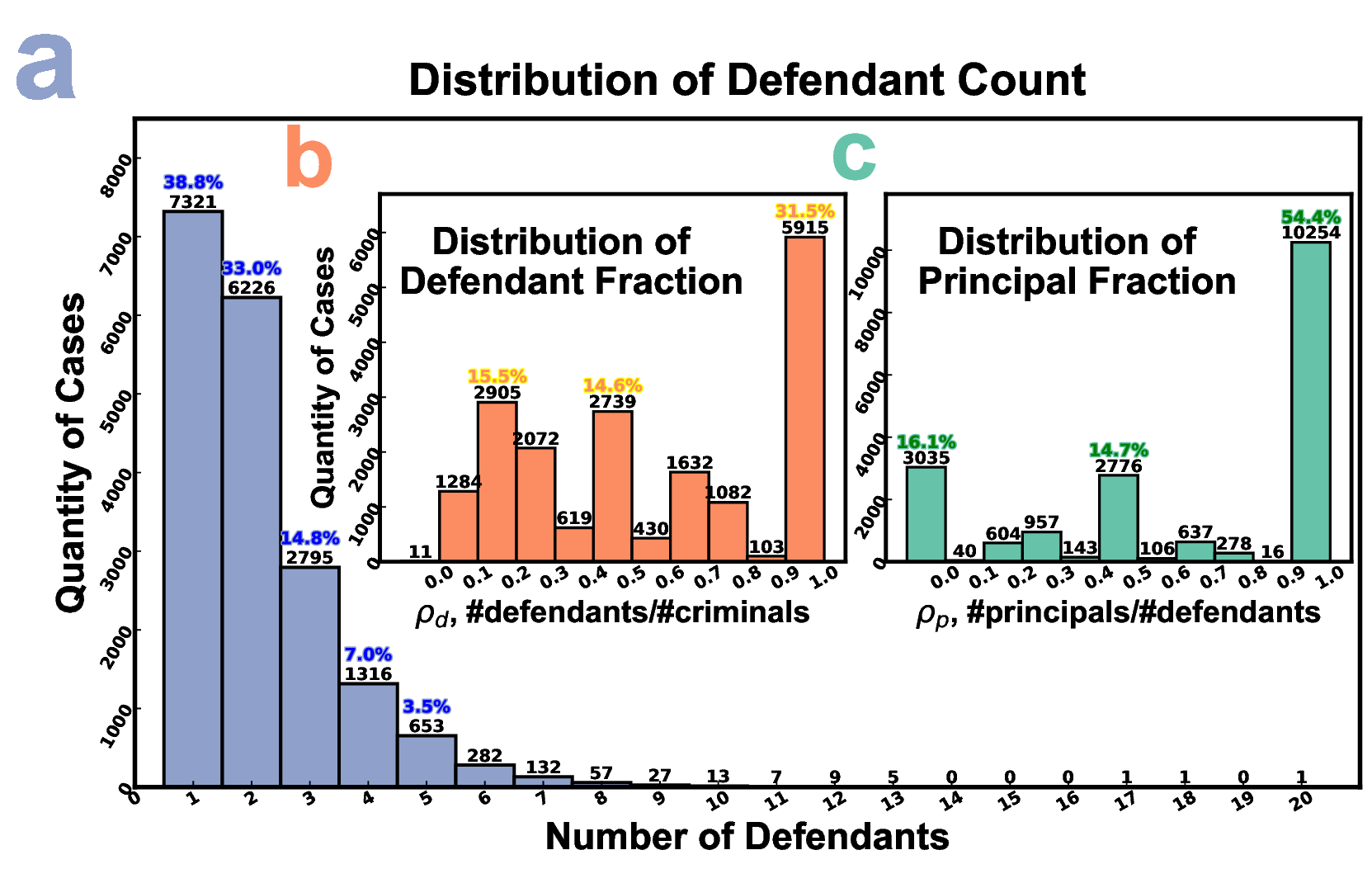}
	\caption{\label{fig:d-d}Defendant distribution in the IMLJP dataset. a) represents the number of defendants per case and their distribution (in blue), b) illustrates the proportion of defendants relative to all individuals involved in each case (in orange), and c) depicts the proportion of defendants identified as principal offenders (in green).}
\end{figure}

In real-world multidefendant cases, suspects are not always jointly charged. Depending on the nature or severity of the crime, prosecutors may choose to indict defendants separately. As a result, judgment documents may contain only verdicts for the defendants involved in the current case, omitting information about other related suspects. Analyzing the behaviors of all suspects within a single judgment document can therefore lead to missing or incomplete data and labels. To explore the distribution of the available DI in multidefendant cases, Fig.~\ref{fig:d-d} illustrates the numbers of defendants and suspects across various cases.

According to Fig. \ref{fig:d-d} (a), a significant portion of multidefendant cases involve only one defendant ($38.8\%$), whereas $58.3\%$ of cases include 2 to 5 defendants. A smaller subset of cases includes more than 5 defendants, with some cases including up to 20 defendants. The defendant distribution map in Fig.~\ref{fig:d-d} (b) indicates that only a limited proportion of the cases involve all the prosecuted defendants ($31.5\%$). Additionally, the data indicate that cases with uniform guilt roles (with either all defendants as principals or all as accessories) account for $70.5\%$ of the total. While cases involving different roles (with both principals and accessories)--offering natural comparisons due to their similar case descriptions--could enhance the comparative learning ability of the model, the limited number of such cases restricts the potential achievable performance improvements (see Fig. \ref{fig:d-d} (c)).

\subsection{Experimental Settings}
\label{sec:experimental_settings}
\textbf{Data selection and partitioning.}
For all model training steps involving the guilt inference model, a comparison dataset comprising both principals and accomplices is curated to ensure high quality and contrastive effectiveness. This dataset contains 6,596 cases involving 20,706 defendants, with 9,464 identified as principals and 11,242 identified as accomplices. To maintain fairness in the training data, equal numbers of principal and accomplice samples are extracted from the category with the fewest labels during repeated experiments. This yields 9,464 pairs of principal and accomplice data obtained through random matching. Given the variability encountered in test verification scenarios, each experiment randomly splits the data into training, validation, and test sets at an 8:1:1 ratio. Furthermore, the training, validation, and test samples used for the experiments comparing effects within the same table are kept consistent.

% \textbf{Parameter setting.}
% Owing to the input limitations of the BERT model, the input length was capped at 512 tokens. During training, we utilized the weighted Adam (AdamW) optimizer \cite{loshchilov2017decoupled}, with a learning rate of \(1 \times 10^{-5}\). The batch size was set to 16, and the maximum number of training epochs for all the models was 10. The loss function for the guilt inference model was defined as the cross-entropy loss, whereas the prison prediction model employed the mean squared error (MSE) as its loss function. The number of resampling iterations was set to 10, with 5 repetitions for each sampling step. All the experiments were conducted on an RTX 3090 graphics card.

\textbf{Parameter setting.}
Owing to the input limitations of BERT-based encoders, the input length is capped at 512 tokens. During training, we use the AdamW optimizer \cite{loshchilov2017decoupled} with a learning rate of \(1 \times 10^{-5}\). The batch size is set to 16, and the maximum number of training epochs for all models is 10. The loss function for the guilt inference model is cross-entropy loss, whereas the prison prediction model employs mean squared error (MSE). The number of resampling iterations is set to 10, with 5 repetitions for each sampling step. All experiments are conducted on an RTX 3090 GPU.

\textbf{Evaluation indicators.}
Since the guilt inference task and the guilt identification task are binary classification problems, we evaluate them using standard metrics: accuracy (Acc.), precision (P.), recall (R.), and F1 score (F1). Given \(N\) samples with true labels \(y_i \in \{0,1\}\) and predictions \(\hat{y}_i \in \{0,1\}\), these metrics are defined as
\[
\text{Acc.} = \frac{1}{N} \sum_{i=1}^N \mathbb{I}(\hat{y}_i = y_i), \quad
\text{P.} = \frac{\sum_{i=1}^N \mathbb{I}(\hat{y}_i = 1 \wedge y_i = 1)}{\sum_{i=1}^N \mathbb{I}(\hat{y}_i = 1)},
\]
\[
\text{R.} = \frac{\sum_{i=1}^N \mathbb{I}(\hat{y}_i = 1 \wedge y_i = 1)}{\sum_{i=1}^N \mathbb{I}(y_i = 1)}, \quad
\text{F1} = \frac{2 \times \text{P.} \times \text{R.}}{\text{P.} + \text{R.}},
\]
where \(\mathbb{I}(\cdot)\) is the indicator function, which equals 1 if the condition inside is true and 0 otherwise. Acc. measures the proportion of correct predictions over all samples. P. indicates how many of the predicted guilt labels are actually correct. R. measures how many of the actual guilty defendants are correctly identified. The F1 is the harmonic mean of precision and recall, balancing both.

The prison prediction task is a regression task, making its measurement standard challenging to define from a judicial perspective. To address this issue, metrics from the China Legal AI Challenge, metrics from prior studies, and self-defined absolute accuracy indicators are included. In this work, the predicted imprisonment value for sample \( i \) is denoted as \(\hat{y}_i\), and the ground truth is denoted as \(y_i\). The employed evaluation metrics are as follows, where $\overline{(*)}$ denotes the mean of $(*)$.

\begin{enumerate}
	\item \textbf{ImpScore} \cite{xiao2018cail, bi2023judicial}. This metric originates from the first China Legal AI Challenge and has been applied in various LJP works. It employs a numerical smoothing technique to confine the prediction accuracy within a range from 0-1 and is defined as \(\text{ImpScore} = \overline{\text{Score1}}\). The definition of \(\text{Score1}\) is as follows:
	\[h_i = \left| \log(\hat{y}_i + 1) - \log(y_i + 1) \right|,\]
	\[
	\text{Score1}_i =
	\begin{cases}
		1, & h_i \leq 0.2, \\
		0.8, & 0.2 < h_i \leq 0.4, \\
		0.6, & 0.4 < h_i \leq 0.6, \\
		0.4, & 0.6 < h_i \leq 0.8, \\
		0.2, & 0.8 < h_i \leq 1, \\
		0, & \text{otherwise}.
	\end{cases}
	\]
	
	\item \textbf{ImpAcc.} This metric is derived from the fourth China Legal AI Challenge and establishes a tolerance range on the basis of various sentencing characteristics, offering a qualitative prediction accuracy assessment. The accuracy is calculated as \(\text{ImpAcc} = \overline{\text{Score2}}\). The definitions for \(\text{Score2}\) are as follows:
	\[
	\text{Score2}_i =
	\begin{cases}
		1, & |y_i - \hat{y}_i| \leq 0.25 y_i, \\
		0, & |y_i - \hat{y}_i| > 0.25 y_i.
	\end{cases}
	\]
	\item \textbf{ImpErr.} To better discern subtle sentencing differences among different defendants in the same case, we adopt an absolute error measure without segmentation functions. To normalize this metric and align the trends of the three metrics, we integrate the maximum detectable prison term into the dataset, which is denoted as \( \text{Prison}_{\text{max}} \). The maximum detected term in our dataset is 180 months; hence, \( \text{Prison}_{\text{max}} = 180 \). The absolute prison prediction error is defined as \(\text{ImpErr} = \overline{\text{Score3}}\), where
	\[
	\text{Score3}_i = \frac{|\hat{y}_i-y_i|}{\text{Prison}_{\text{max}}}.
	\]
\end{enumerate}

Among all the evaluation metrics mentioned above, higher ImpScore and ImpAcc values indicate that the predicted values are closer to the ground truths, whereas ImpErr, as a measure of error, indicates closer proximity to the ground truth when it is smaller.

\subsection{Baselines}
To comprehensively evaluate the proposed MMSI framework, we compare it against a wide range of baseline methods spanning traditional single-defendant frameworks, prompt-based generative models, pretrained legal-domain models, and SOTA LLMs. The baseline models are grouped into four categories:

\begin{itemize}
    \item \textbf{Single-defendant judgment prediction methods.}
    \begin{itemize}
        \item \textbf{NeurJudge} \cite{yue2021neurjudge}: A pipeline-based model that uses intermediate subtask outputs to enhance final judgment predictions. 
        \item \textbf{LADAN} \cite{xu2020distinguish}: A graph neural network approach that captures fine-grained distinctions among legal statutes.
    \end{itemize}

    \item \textbf{Prompt-based generative models.}
    \begin{itemize}
        \item \textbf{mT5} \cite{xue2020mt5}: A multilingual encoder-decoder model adapted to the legal domain using prompt tuning for generation-based sentence prediction.
        \item \textbf{Finetune-CoT} \cite{ho2024large}: A fine-tuned model leveraging chain-of-thought style prompts to enhance sentence reasoning and explainability.
        \item \textbf{HRN} \cite{lyu2023multi}: A hierarchical reasoning framework for multidefendant sentencing, which models the legal process as a stepwise generative reasoning chain. This is the only available multidefendant-specific public baseline.
    \end{itemize}

    \item \textbf{Legal-domain models with rRegression adaptation.}
    \begin{itemize}
        \item \textbf{Legal-BERT} \cite{chalkidis2020legal} and \textbf{Lawformer} \cite{xiao2021lawformer}: Originally designed for classification tasks, these models are adapted for regression by replacing their classification heads with a regression layer. We retain their native tokenizers and encoders.
    \end{itemize}
    
    \item \textbf{SOTA LLMs.}
    \begin{itemize}
        \item \textbf{GPT-3.5}, \textbf{GPT-4o} \footnote{\url{https://chatgpt.com/}}, \textbf{LLaMA3-70B} \footnote{\url{https://www.llama.com/}}, \textbf{Claude-3} \footnote{\url{https://www.anthropic.com/claude}}, \textbf{Gemini-1.5} \footnote{\url{https://gemini.google.com/}}, and \textbf{DeepSeek-V3} \footnote{\url{https://www.deepseek.com/}}: These models are evaluated using prompt-based queries via public APIs. They represent current SOTA capabilities in general-purpose LLM-based legal reasoning.
    \end{itemize}
\end{itemize}

In addition to these baselines, we report several variants of the proposed MMSI framework for comparison. The default MMSI model uses the BERT-base-Chinese encoder as its backbone under a unified BERT-base-Chinese tokenizer and embedding front end. The alternative variants differ only in the choice of encoder, while all other components are kept fixed:
\begin{itemize}
\item \textbf{$\text{MMSI}_{\text{Legal-BERT}}$}: 
Replace the encoder in the default MMSI with Legal-BERT \cite{chalkidis2020legal}, which shares the BERT architecture but is pretrained on legal-domain corpora.

\item \textbf{$\text{MMSI}_{\text{Lawformer}}$}: 
Replace the encoder in the default MMSI with Lawformer \cite{xiao2021lawformer}, which modifies the BERT architecture for long-text modeling.

\item \textbf{$\text{MMSI}_{\text{Muppet}}$}: 
Replace the encoder in the default MMSI with Muppet \cite{armen2021DBLPmuppet}, a BERT-like encoder pretrained with multitask objectives on diverse NLP data.
\end{itemize}

Furthermore, to improve the proposed technique, we conduct a multi-task joint training experiment based on the BERT architecture. The resulting model, denoted as $\text{MMSI}^{\text{Inject}}$, is reported for comparison. Detailed experimental settings and analyses are provided in Appendix~\ref{appsec:E_joint_MMSI}. Prompt-based generative models are fine-tuned on the same training data, while API-accessed LLMs are evaluated in zero-shot settings. To ensure fairness, we adopt a standardized prompt format (see Appendix~\ref{sec:K_prompt_info}, referenced in Appendix~\ref{appsec:F_additional_experiments}).

\begin{table*}[ht]
	\centering
	\caption{Comparison among the sentencing results produced by the existing and proposed methods across various models.}
	\label{tab:Baseline}
	\begin{tabular}{l|lccc}
		\hline
        \textbf{Category} & \textbf{Method} & \textbf{ImpScore ($\pm$95\%CI)} & \textbf{ImpAcc ($\pm$95\%CI)} & \textbf{ImpErr ($\pm$95\%CI)}\\
		\hline
        \multirow{2}{*}{Single-defendant} 
            & LADAN & 0.2878 ($\pm$0.0073) & 0.1482 ($\pm$0.0053) & 0.1519 ($\pm$0.0008) \\
            & NeurJudge & 0.4327 ($\pm$0.0137) & 0.2432 ($\pm$0.0112) & 0.1398 ($\pm$0.0052) \\
        \hline
        \multirow{3}{*}{Prompt-based} 
            & mT5 & 0.3917 ($\pm$0.0255) & 0.2317 ($\pm$0.0261) & 0.1558 ($\pm$0.0106) \\
            & Finetune-CoT & 0.4110 ($\pm$0.0080) & 0.2237 ($\pm$0.0081) & 0.1384 ($\pm$0.0031) \\
            & HRN & 0.4884 ($\pm$0.0212) & 0.1961 ($\pm$0.0251) & 0.1245 ($\pm$0.0118) \\
        \hline
        \multirow{2}{*}{Regression-adapted} 
            & Legal-BERT & 0.4382 ($\pm$0.0034) & 0.2126 ($\pm$0.0028) & 0.1307 ($\pm$0.0018) \\
            & Lawformer & 0.4816 ($\pm$0.0572) & 0.2694 ($\pm$0.0454) & 0.1236 ($\pm$0.0103) \\
        \hline
        \multirow{6}{*}{SOTA LLMs} 
            & Claude-3 & 0.3570 ($\pm$0.0235) & 0.1693 ($\pm$0.0232) & 0.3721 ($\pm$0.3451) \\
            & GPT-3.5 & 0.5527 ($\pm$0.0072) & 0.2719 ($\pm$0.0088) & 0.1223 ($\pm$0.0035) \\
            & LLaMA3-70B & 0.5798 ($\pm$0.0073) & 0.2898 ($\pm$0.0091) & 0.1085 ($\pm$0.0118) \\
            & Gemini-1.5 & 0.5979 ($\pm$0.0073) & 0.3250 ($\pm$0.0098) & 0.1047 ($\pm$0.0107) \\
            & GPT-4o & 0.6271 ($\pm$0.0007) & 0.2977 ($\pm$0.0090)& 0.1185 ($\pm$0.0003) \\
            & DeepSeek-V3 & 0.7543 ($\pm$0.0052) & 0.4505 ($\pm$0.0097) & 0.0655 ($\pm$0.0021) \\
		\hline
        \multirow{4}{*}{MMSI w/ alt encoders} 
            & $\text{MMSI}_{\text{Lawformer}}$ & 0.7374 ($\pm$0.0397) & 0.4746 ($\pm$0.0330) & 0.0709 ($\pm$0.0132) \\
            & $\text{MMSI}$ & 0.7526 ($\pm$0.0119) & 0.4653 ($\pm$0.0160) & 0.0680 ($\pm$0.0027) \\
            & $\text{MMSI}_{\text{Legal-BERT}}$ & 0.7570 ($\pm$0.0234) & 0.4770 ($\pm$0.0277) & 0.0641 ($\pm$0.0047) \\
            & $\text{MMSI}_{\text{Muppet}}$ & \textbf{0.7851} ($\pm$\textbf{0.0110}) & \textbf{0.5083} ($\pm$\textbf{0.0152}) & \textbf{0.0607} ($\pm$\textbf{0.0022}) \\
        \hline
        \multirow{1}{*}{Joint-training} 
            & $\text{MMSI}^{\text{Inject}}$ & 0.7826 ($\pm$0.0150) & 0.4979 ($\pm$0.0166) & 0.0658 ($\pm$0.0061) \\
        \hline
        \multicolumn{5}{r}{\footnotesize{$\pm$95\% CI: 95\% confidence interval}}\\
	\end{tabular}
\end{table*}

\begin{figure*}[!t]
	\centering
	\includegraphics[width=0.8\linewidth]{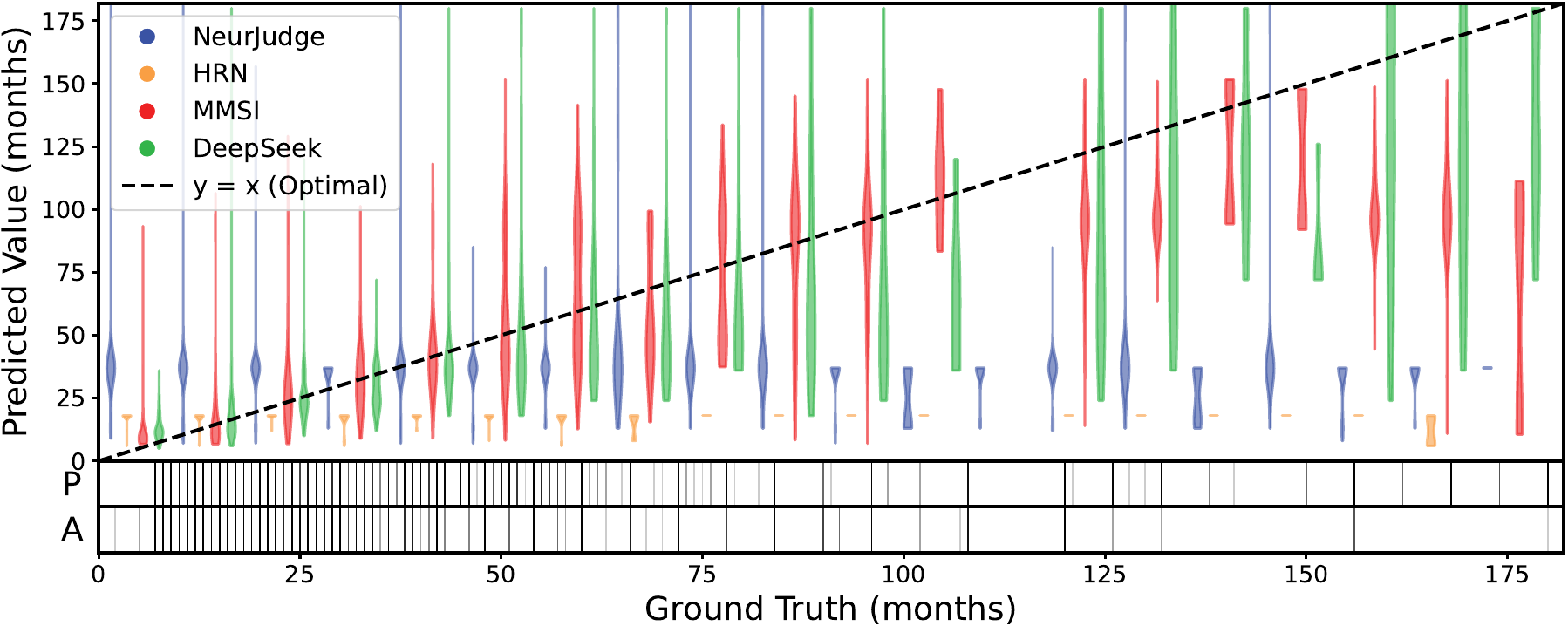}
	\caption{\label{fig:baseline_visible}Visualization of the prediction results yielded by the four approaches. The two vertical line plots below represent the distributions of the true prison term values (P for principal, A for accomplice), where darker colors indicate denser distributions. The approaches compared include the FD-based NeurJudge (blue), the multidefendant HRN (orange), the proposed MMSI (red). and the SOTA LLM DeepSeek (green). An optimal dashed line represents the alignment between ground truth and predictions, highlighting the prediction accuracy of each method.}
\end{figure*}

The experimental results are summarized in Table~\ref{tab:Baseline}. 
Among non-LLM baselines, MMSI outperforms all baseline models across all three evaluation metrics. 
Notably, the default MMSI model achieves performance comparable to the best-performing SOTA LLM (DeepSeek-V3). 
When we replace its encoder with alternative pretrained encoders, the variants $\text{MMSI}_{\text{Legal-BERT}}$ and $\text{MMSI}_{\text{Muppet}}$ further surpass DeepSeek-V3, with $\text{MMSI}_{\text{Muppet}}$ achieving the best overall results. 
Moreover, the jointly trained multi-task $\text{MMSI}^{\text{Inject}}$ model achieves higher ImpScore and ImpAcc than all SOTA LLMs.
These results highlight the effectiveness of MMSI in handling complex multidefendant sentencing tasks by jointly modeling guilt inference and sentencing. 
Additional experiments provided in Appendix~\ref{appsec:F_additional_experiments} further demonstrate the effectiveness of the MASK strategy and label injection method.

To ensure a fair and representative comparison, we select NeurJudge, HRN, and DeepSeek-V3 as the best-performing representatives from the aforementioned three categories of baseline methods. We compare their sentencing predictions with those of MMSI and visualize the distribution of true versus predicted prison terms, as shown in Fig.~\ref{fig:baseline_visible}. Among the compared models, the classification-based NeurJudge concentrates predictions within narrow ranges, while the MMSI, which adopts a regression approach, demonstrates greater adaptability to the full spectrum of sentencing outcomes. The lower performance of HRN stems from its data-driven generative framework, which tends to cluster predictions around common prison-term lengths that occur with high frequency in the training data, limiting its capacity to distinguish between roles, as evident from the visualization. Compared with DeepSeek-V3, MMSI exhibits superior performance in mid-range prison predictions. However, performance deteriorates on cases with longer sentences. This limitation arises because MMSI relies primarily on principal/accomplice role information inferred from the FD, whereas DeepSeek-V3 has access to a broader spectrum of contextual features from the full FD. This suggests that additional sentencing-related cues embedded in the FD remain underutilized by MMSI and offer a promising direction for future improvements.

\begin{figure}[!ht]
	\centering
	\includegraphics[width=0.95\linewidth]{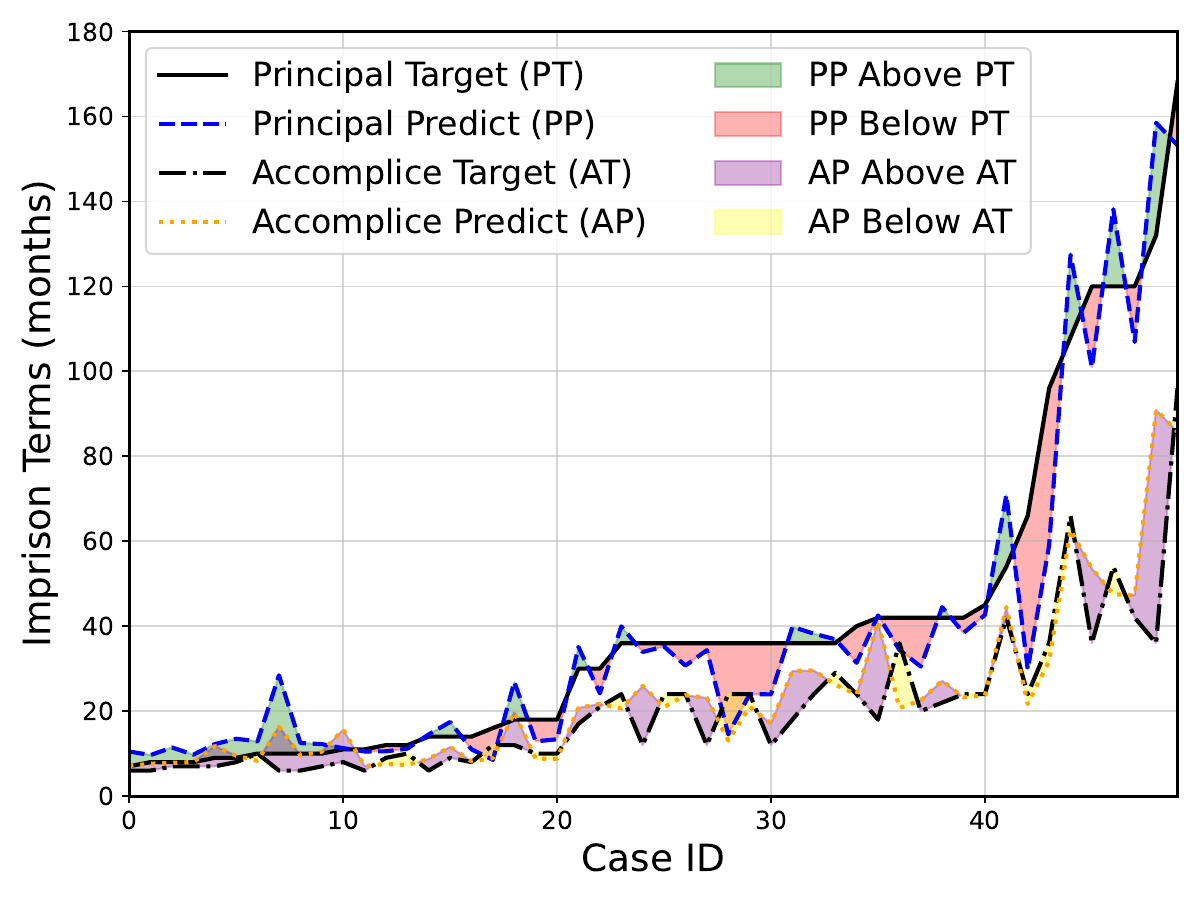}
	\caption{\label{fig:MMSIcaseidcompare}Visualization of the ability to differentiate between principal and accomplice defendants in multidefendant cases. The x-axis represents individual cases, whereas the y-axis denotes prison term values. The ground-truth values are marked as a solid black line for principals (PT) and a dashed black line for accomplices (AT). The predictions are indicated as a dashed blue line for principals (PP) and a dashed orange line for accomplices (AP). To highlight the relationship between the predicted and actual values, shaded regions illustrate the observed discrepancies: green for \(PP > PT \), pink for \(PP < PT \), purple for \(AP > AT \), and yellow for \(AP < AT \).}
\end{figure}

Most trained baseline models tend to predict common sentence lengths within each offense type, limiting their ability to capture the full range of sentencing outcomes. In contrast, the single-sentence prediction approach of MMSI better identifies key sentencing factors, enabling clearer differentiation between the principal and accomplice roles. To demonstrate the ability of MMSI to differentiate between the principal and accomplice roles, we randomly select cases involving both roles. Fig.~\ref{fig:MMSIcaseidcompare} illustrates the relationship between the true and predicted sentence values for each role. The results clearly show that, in real-world multidefendant cases, there is indeed a sentencing disparity between principals and accomplices, with principals receiving longer sentences than accomplices. MMSI effectively adjusts its predictions based on the context of each case. When true sentence values are moderate, the model effectively captures the distinct responsibilities of each role, providing clear and accurate predictions. Overall, MMSI demonstrates strong defendant-level role differentiation capabilities and delivers precise sentencing recommendations in multidefendant cases.

\subsection{Interpretability Analysis}
To further investigate the internal mechanisms of the proposed model and its ability to distinguish multiple defendants, we adopt Integrated Gradients (IG) \cite{sundararajan2017axiomatic} to quantify the contribution of each input token to the model’s predictions. This token-level analysis highlights legally relevant phrases and aligns model attributions with judicial factors.

\begin{figure*}[!t]
    \centering
    \begin{subfigure}[b]{0.8\linewidth}
        \centering
        \includegraphics[width=\linewidth]{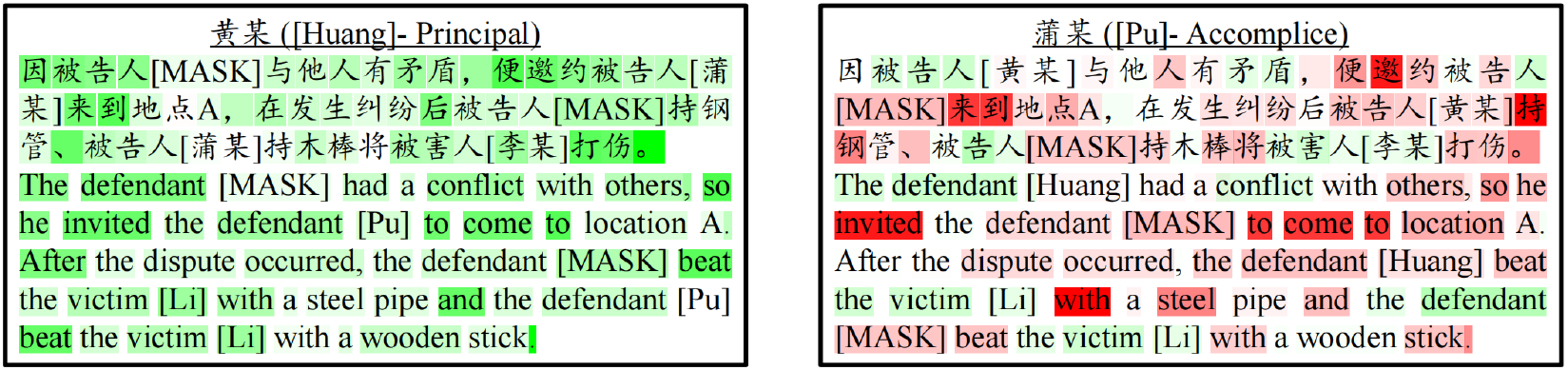}
        \caption{}
        \label{fig:case_study_a}
    \end{subfigure}

    \vspace{-1em}
    \noindent\makebox[\textwidth]{\dotfill} 
    \vspace{-0.8em}

    \begin{subfigure}[b]{0.8\linewidth}
        \centering
        \includegraphics[width=\linewidth]{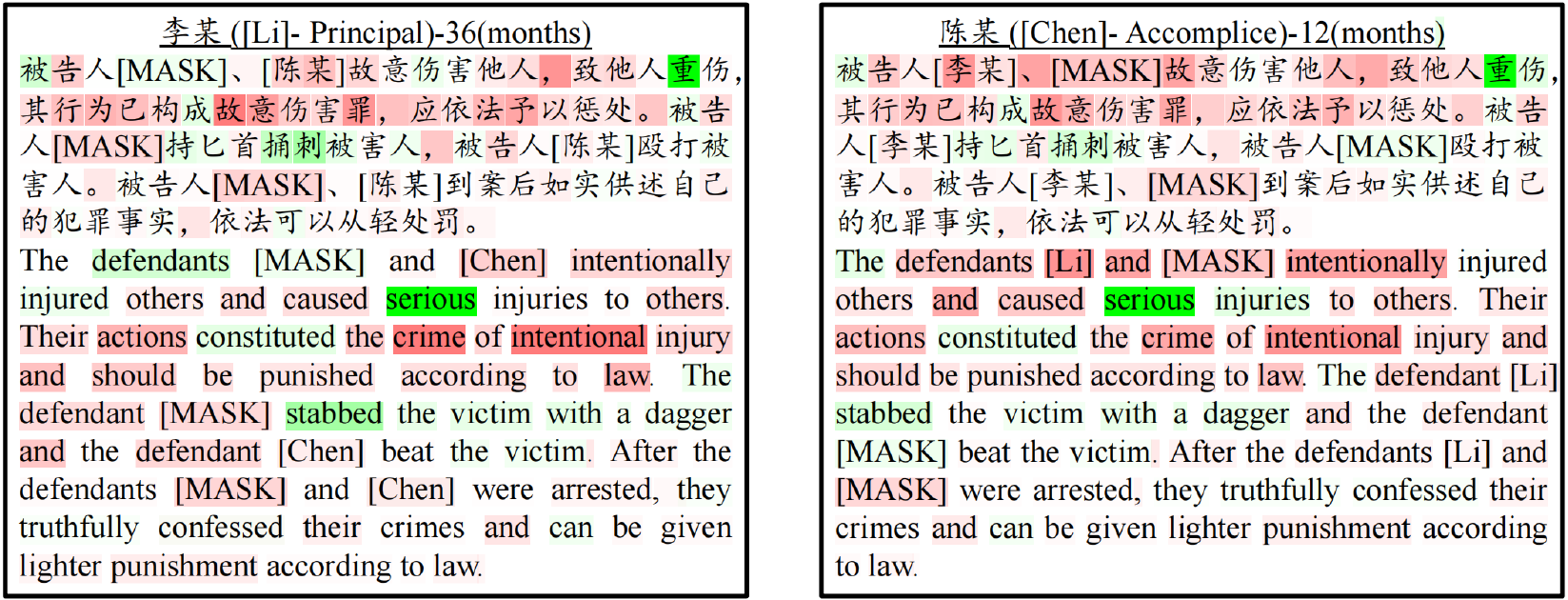}
        \caption{}
        \label{fig:prison_study}
    \end{subfigure}

    \caption{IG-based visualization of token attributions for two key tasks. 
    Subfigure (a) shows results for the guilt inference model, and subfigure (b) for the prison prediction model. 
    Token-level contributions are computed using the IG method following the Captum interpretability framework.
    Highlighting uses a green-to-red color gradient, where deeper green denotes stronger attribution toward predicting the principal role. Masked defendant names, role labels, and imprisonment terms are shown in the first row.\label{fig:intepret}}
\end{figure*}

We present two representative cases to assess MMSI’s interpretability in guilt inference and prison prediction tasks. As illustrated in Fig.~\ref{fig:case_study_a}, both inputs share near-identical wording, with the only variation being the entity represented by [MASK]. When [MASK] refers to the principal offender, MMSI assigns high attribution scores to action-related tokens (e.g., “invited”, “beat”), highlighting its recognition of primary culpability. In contrast, when [MASK] indicates an accomplice, attribution scores for the same tokens drop markedly. This contrast reveals the model’s ability to differentiate roles based on subtle contextual cues.

Fig.~\ref{fig:prison_study} demonstrates similar behavior in the sentencing task. Although the narrative structure remains consistent, the defendant’s involvement differs. When [MASK] performs the stabbing, tokens such as “stabded” and “serious” receive high attribution, leading to a longer predicted sentence. When [MASK] only participates in beating, the attribution to key legal tokens weakens, resulting in a shorter sentence. These observations highlight MMSI’s capacity to infer outcome-relevant distinctions under minimal input variation. Furthermore, additional evidence supporting the model’s interpretability is provided through attention visualization in Appendix~\ref{appsec:G_attention_case_study}.

To further evaluate interpretability, we adopt comprehensiveness (COMP) \cite{deyoung2019eraser}, measuring performance drops after removing top-$k$ influential tokens:
\begin{align*}
\text{COMP}_M(k) &= M(x) - M(\tilde{x}^{(k)}), \\
\text{COMP}_p(k) &= P(x)_y - P(\tilde{x}^{(k)})_y,
\end{align*}
where $M$ denotes a regression metric (e.g., ImpScore, ImpAcc, or ImpErr), and $P(x)_y$ is the predicted probability for the correct class in the classification task. Here, $\tilde{x}^{(k)}$ denotes the perturbed input obtained from $x$ after removing the top-$k$ influential tokens.

Note that while larger COMP values indicate stronger feature importance for metrics such as ImpScore and ImpAcc (where higher values are better), some regression metrics like ImpErr are minimized (i.e., lower values are better). For these, the direction of COMP should be inverted or interpreted accordingly to maintain consistent meaning.

\begin{table}[!htbp]
\centering
\caption{Comprehensiveness (COMP) scores for top-$k$ token removal.}
\label{tab:comp_scores}
\resizebox{\columnwidth}{!}{
\begin{tabular}{c|ccc|c}
\hline
\textbf{Top-k} & \textbf{COMP$_{\text{ImpScore}}$} & \textbf{COMP$_{\text{ImpAcc}}$} & \textbf{COMP$_{\text{ImpErr}}$} & \textbf{COMP$_p$}\\
\hline
1 & +0.1651 & +0.1782 & -0.0239 & +0.0284\\
2 & +0.2446 & +0.2327 & -0.0546 & +0.0767\\
3 & +0.2572 & +0.2436 & -0.0581 & +0.1294\\
4 & +0.2677 & +0.2505 & -0.0613 & +0.1576\\
5 & +0.2735 & +0.2545 & -0.0630 & +0.1627\\
\hline
\end{tabular}
}
\end{table}

As presented in Table~\ref{tab:comp_scores}, removing top-$k$ tokens results in consistent performance degradation, confirming that the attributed features are both informative and legally meaningful. Together, these qualitative and quantitative analyses demonstrate the interpretability and contextual sensitivity of MMSI. Additionally, we provide supplementary explanations in Appendix~\ref{appsec:H_word_cloud}, including word clouds of the top-5 important tokens for the regression model, to further support interpretability. The results indicate that the model effectively focuses on key tokens related to severe outcomes (e.g., “serious injury,” “death”), collective criminal behavior (e.g., “assault”), and weapons (e.g., “knife”), while the [MASK] tokens play a crucial role in emphasizing the behavioral roles of individual defendants.

%--- Section ---%
\section{Conclusion and future work}
\label{sec:conclusions}
This work addresses the prediction of guilt and sentencing in multidefendant criminal cases. An analysis of judicial logic reveals that a defendant's role, whether they are a principal or an accomplice, significantly impacts their sentencing outcome. However, the intertwined nature of judgment documents complicates the task of clearly differentiating among the roles of defendants. To address this issue, we introduce an oriented defendant role masking technique, which proves effective in both guilt identification and inference tasks. To further enhance the role differentiation effect of the model, we develop a comparative data construction method that leverages natural contrasts between roles within the same case, yielding superior predictive outcomes to those obtained with the random data construction scheme. Additionally, by leveraging judicial logic, guilt labels are propagated and integrated into the final sentencing regression model, effectively addressing input-length constraints. This culminates in the MMSI method for multidefendant sentencing prediction, which encodes legal inference patterns in a multistage neural framework to ensure interpretability grounded in legal logic. Utilizing intentional injury cases as examples, we construct a dataset, named IMLJP, comprising 6,596 multidefendant cases for analysis purposes. The results obtained on this real-world dataset confirm the effectiveness and enhanced performance of our proposed approach, surpassing SOTA LLMs. Furthermore, MMSI also achieves outstanding results on the multidefendant dataset CMDL. Additional ablation and sensitivity analyses, reported in Appendix~\ref{appsec:I_ablation} and Appendix~\ref{appsec:J_sensitivity}, revealing that the proposed model exhibits strong capabilities in prison prediction. These improvements are attributed to the masking strategy and the enhanced model representations derived from label propagation. MMSI also demonstrates notable robustness across different hyperparameter configurations. Moreover, the multistage reasoning paradigm of MMSI can be transferred to the reasoning chains of LLMs, and the results in Appendix~\ref{appsec:L_multi-stage_prompt} further confirm that such multistage, emphasis-driven strategies can improve LLM performance.

This work opens avenues for further exploration in several ways. First, the oriented masking technique used for prison prediction is not limited to multidefendant cases; it can also enhance single-defendant analyses by allowing the model to focus more on the contextual information concerning defendants. Second, while judgment prediction ideally encompasses all final court verdicts, it should include prison and probation predictions. Probation considerations, such as whether to suspend a sentence and the duration of probation, involve factors such as social impacts and remorse, which differ from the core issues of multidefendant guilt. Thus, this work suggests extending the framework into these areas. The decision to suspend a sentence represents a binary prediction, whereas selecting the probation term is a regression task, both of which are consistent with the core algorithms addressed in this work for guilt inference and sentence prediction. Additionally, the developed dataset already includes labels for these tasks, offering opportunities for further research. Corresponding tasks can be designed to utilize classification models for predicting probation eligibility and regression models for estimating probation terms. In these models, the legal elements of probation can guide the process of designing appropriate label propagation mechanisms to achieve enhanced predictive capabilities. Moreover, although this work focuses on Chinese intentional injury cases, the proposed MMSI framework is conceptually generalizable. Many other legal systems, including civil law jurisdictions (e.g., Germany, France, Japan) and common law jurisdictions (e.g., the United States and the United Kingdom), also differentiate sentencing based on the roles of defendants, such as principal versus accomplice. The oriented masking mechanism can flexibly encode role-related placeholders across these systems, while the guilt label broadcasting strategy can be extended to multilingual corpora using multilingual pretrained models. These features suggest strong potential for cross-jurisdictional and cross-lingual adaptation. Finally, owing to data limitations, this work focuses on intentional injury cases, but future efforts will aim to extend the guilt inference and sentencing prediction tasks to other types of crimes involving multiple defendants.

This work examines judicial sentencing logic by integrating legal frameworks with data-driven analysis and proposes an intelligent solution for completing guilt inference and sentencing prediction tasks in multidefendant cases. The proposed approach offers a practical foundation for developing advanced auxiliary judicial tools.

\begin{appendices}
\label{appsec:si}

\section{Comparison of Tokenizer-Encoder Models for Legal Text Encoding}
\label{appsec:A_tokenizer}
Since the available data consist of text, they must be converted into machine-readable encodings for processing, making natural language encoding an essential preprocessing step in this work. Various pretrained language models were selected for comparison on the basis of their linguistic characteristics and relevance, including BERT (BERT-base-Chinese and BERT-base-uncased \cite{jacob2018DBLPchinese}), RoBERTa (RoBERTa-base \cite{liu2019DBLProberta}), Muppet (Muppet-RoBERTa-base \cite{armen2021DBLPmuppet}), and the Chinese legal-domain model Lawformer \cite{xiao2021lawformer}. This comparative analysis focuses on guilt inference with FDs as the primary evaluation task. The results are presented in Table \ref{tab:model_compare_tasks}.

\begin{table*}[!t]
	\centering
	\caption{Comparison of different tokenizer-encoder models on guilt inference.}
	\label{tab:model_compare_tasks}
	\begin{tabular}{l|cccc}
		\hline
		Models & Acc.($\pm$95\%CI) & P.($\pm$95\%CI) & R.($\pm$95\%CI) & F1.($\pm$95\%CI) \\
		% & (95\%CI) & (95\%CI) & (95\%CI) & (95\%CI) \\
		\hline
		BERT-base-Chinese & \textbf{0.8773} ($\pm$0.0028) & \textbf{0.8669} ($\pm$0.0044) & \textbf{0.8856} ($\pm$0.0043)& \textbf{0.8760} ($\pm$0.0028)\\
		% \hline
		BERT-base-uncased & 0.8361 ($\pm$0.0029) & 0.8396 ($\pm$0.0040) & 0.8339 ($\pm$0.0038)& 0.8366 ($\pm$0.0029)\\
		% \hline
		RoBERTa-base & 0.7152 ($\pm$0.0455)& 0.7059 ($\pm$0.0913)& 0.6508 ($\pm$0.0907)& 0.6624 ($\pm$0.0851)\\
		% \hline
		Muppet-RoBERTa-base & 0.8335 ($\pm$0.0032)& 0.7930 ($\pm$0.0053)& 0.8632 ($\pm$0.0046)& 0.8264 ($\pm$0.0035)\\
		% \hline
		RoBERTa-Lawformer & 0.6752 ($\pm$0.0318) & 0.6884 ($\pm$0.0758)& 0.6441 ($\pm$0.0692)& 0.6463 ($\pm$0.0630)\\
		\hline
        \multicolumn{5}{r}{\footnotesize{$\pm$95\% CI: 95\% confidence interval}}\\
	\end{tabular}
\end{table*}

\begin{table*}[!htbp]
	\centering
	\caption{Evaluation of dataset construction approaches for guilt inference (FD-Based) and identification (CV-Based) tasks}
	\label{tab:train_methods_result}
	\begin{tabular}{llcccc}
		\toprule
		\multirow{2}{*}{Task} & \multirow{2}{*}{Method} & \multicolumn{4}{c}{Metrics (Mean $\pm$ 95\%CI)} \\
		\cmidrule(lr){3-6}
		& & Acc. & P. & R. & F1 \\
		\midrule
		\multirow{3}{*}{FD-Based} & Pairs & \textbf{0.8766} ($\pm$0.0032) & \textbf{0.8662} ($\pm$0.0041) & \textbf{0.8849} ($\pm$0.0047) & \textbf{0.8753} ($\pm$0.0031) \\
		& Random & 0.7977 ($\pm$0.0041) & 0.8539 ($\pm$0.0052) & 0.7681 ($\pm$0.0055) & 0.8085 ($\pm$0.0035) \\
		& Full & 0.8050 ($\pm$0.0039) & 0.9129 ($\pm$0.0039) & 0.7511 ($\pm$0.0042) & 0.8241 ($\pm$0.0032) \\
		\midrule
		\multirow{3}{*}{CV-Based} & Pairs & \textbf{0.9925} ($\pm$0.0010) & \textbf{0.9915} ($\pm$0.0014) & \textbf{0.9935} ($\pm$0.0017) & \textbf{0.9925} ($\pm$0.0010) \\
		& Random & 0.9833 ($\pm$0.0017) & 0.9847 ($\pm$0.0024) & 0.9820 ($\pm$0.0026) & 0.9834 ($\pm$0.0017) \\
		& Full & 0.9881 ($\pm$0.0013) & 0.9943 ($\pm$0.0012) & 0.9821 ($\pm$0.0026) & 0.9881 ($\pm$0.0013) \\
		\bottomrule
		\multicolumn{6}{r}{\footnotesize{$\pm$95\% CI: 95\% confidence interval}}\\
	\end{tabular}
	%}
\end{table*}

\begin{table*}[!htbp]
	\centering
	\caption{Oriented MASK effectiveness comparison in guilt identification tasks.}
	\label{tab:MASK_effect_2}
	\begin{tabular}{l|cccc}
		\hline
		Method & Acc. ($\pm$95\%CI) & P. ($\pm$95\%CI) & R. ($\pm$95\%CI) & F1. ($\pm$95\%CI)\\
		\hline
		Original & 0.5000 ($\pm$0.0000) & 0.4885 ($\pm$0.1188) & 0.4100 ($\pm$0.0551) & 0.3808 ($\pm$0.0801)\\
		\hline
		Split & 0.6823 ($\pm$0.0062) & 0.6217 ($\pm$0.0124)& 0.7096 ($\pm$0.0097)& 0.6613 ($\pm$0.0073)\\
		\hline
		MASK & \textbf{0.9904} \textbf{($\pm$0.0010)} & \textbf{0.9899} \textbf{($\pm$0.0016)} & \textbf{0.9908} \textbf{($\pm$0.0012)} & \textbf{0.9903} \textbf{($\pm$0.0010)} \\
		\hline
        \multicolumn{5}{r}{\footnotesize{$\pm$95\% CI: 95\% confidence interval}}\\
	\end{tabular}
\end{table*}

Results indicates that BERT-base-Chinese outperformed all other models across tasks, leading to its adoption as the foundation for subsequent experiments. 

\section{Data Construction and Validation Comparisons}
\label{appsec:B_consdata}
In multidefendant cases, classification models often struggle to distinguish the expressions and contexts associated with different roles, such as principals and accomplices. The oriented masking method aids in the recognition task by focusing on the target defendant, but further enhancing the sensitivity of the model to subtle role distinctions presents a new challenge. Since these roles are frequently contrasted within the same text, introducing similar yet distinctly labeled data can improve the ability of the model to detect these nuanced differences. Therefore, this section examined various dataset construction methods tailored for two main tasks: guilt inference based on FDs and guilt identification based on CVs. These tailored methods aim to increase the effectiveness of the model in handling the complexities of role differentiation tasks in multidefendant cases. The following data construction strategies were evaluated.
\begin{enumerate}
	\item Pairs dataset (\textbf{Pairs}): Cases with both principals and accomplices were screened, extracting masked data from both roles within the same case to create representative comparative training pairs.
	\item Random dataset (\textbf{Random}): Principal and accomplice data were randomly selected, ensuring equal numbers of positive and negative samples to match the paired approach.
	\item Full dataset (\textbf{Full}): Principal and accomplice data were extracted while balancing the numbers of positive and negative samples based on the smaller available counts.
\end{enumerate}

Importantly, while the Pairs and Random datasets had the same sample size, the Pairs dataset specifically included opposing samples, i.e., those with different labels that closely resembled their original expressions or embeddings. The Full dataset served as an expansion of the Random dataset. Intuitively, increasing the data volume can enhance the prediction accuracy to some extent and can also provide coverage for the Pairs dataset with a certain probability. The corresponding results are summarized in Table \ref{tab:train_methods_result}.

Results demonstrate that the Pairs training data significantly enhanced the ability of the model to distinguish between the semantic features of different individuals within the same expression. The accuracy achieved with the Pairs dataset was notably higher than that achieved with the Random dataset. Although increasing the size of the Random dataset led to some accuracy improvement, its performance still lagged behind that of the Pairs dataset. This highlights the effectiveness of the paired data construction method introduced in this work, indicating the considerable potential for attaining further performance improvements with the inclusion of additional data in the future.

\section{Additional experiments involving masking strategies}
\label{appsec:C_extr_FD_exp}
This section further presents the application effects provided by the oriented defendant name masking method in the guilt identification task implemented based on CVs. In a similar manner to the guilt inference task based on FDs, the performance of the Original, Split, and MASK methods was evaluated. The results are summarized in Table \ref{tab:MASK_effect_2}.

The results indicate that the oriented masking mechanism demonstrated excellent performance in the guilt identification task. Notably, it achieves exceptional performance in the guilt classification task. This is expected, as standard judgment documents often include explicit keywords such as “principal” or “accomplice” immediately following the target defendant's name, allowing simple rule-based extraction methods to yield high accuracy. As such, this CV-based task serves primarily as a validation case to demonstrate the effectiveness of our masking strategy. In contrast, the core focus of this work remains on guilt inference based on FDs, where such explicit lexical cues are absent and rule-based methods become ineffective.

\section{Transfer Evaluation Using CMDL}
\label{appsec:D_cmdl}
To evaluate the generalizability of the proposed MMSI method, we conduct comparative experiments on the CMDL dataset \cite{huang2024cmdl}, using two training setups with different data scales (small-scale and big-scale). Since our current MMSI implementation is tailored specifically for the crime of intentional injury, we filter the CMDL dataset to retain only cases related to this charge for training, validation, and testing.
\begin{table*}[h]
	\centering
	\caption{Performance of MMSI and baseline models on the CMDL dataset under zero-shot and training-based settings.}
	\label{tab:cmdl}
	\begin{tabular}{lcccc}
		\toprule
		Method &Dataset& ImpScore ($\pm$95\%CI)& ImpAcc ($\pm$95\%CI)& ImpErr ($\pm$95\%CI)\\
		\midrule
		\multirow{2}{*}{mT5-T} & small & 0.5176 ($\pm$0.0485) &0.2694 ($\pm$0.0559) & 0.0963 ($\pm$0.0206)\\
		&big & 0.4007 ($\pm$0.0246) &0.2257 ($\pm$0.0251)& 0.1336 ($\pm$0.0083)\\
		\midrule
		\multirow{2}{*}{MMSI-P} & small & 0.7142 ($\pm$0.1645) & 0.3979 ($\pm$0.1578)& 0.0432 ($\pm$0.0182) \\
		& big & 0.6994 ($\pm$0.1520) &  0.3675 ($\pm$0.1415)& 0.0495 ($\pm$0.0163) \\
		\midrule
		\multirow{2}{*}{MMSI-T} & small & 0.7300 ($\pm$0.0146) & 0.4343 ($\pm$0.0296)& 0.0539 ($\pm$0.0025) \\
		& big & 0.6485 ($\pm$0.0634) &  0.3767 ($\pm$0.0492)& 0.0694 ($\pm$0.0026) \\
		\bottomrule
		\multicolumn{5}{r}{\footnotesize{$\pm$95\% CI: 95\% confidence interval}}\\
	\end{tabular}
\end{table*}

In addition, the CMDL dataset lacks complete annotation of principal/accomplice roles required by MMSI. To address this, we apply a data filtering strategy that retains only cases in which the CV explicitly mentions the keywords principal or accomplice. For cases without such labels in the training set, we use our previously validated guilt identification model (achieving 99\% accuracy) to automatically reconstruct role annotations from the CVs, enabling consistent supervision without manual labeling.

We compare the following models:
\begin{itemize}
    \item mT5-T: The mT5 model trained under the small and big training setups.
    \item MMSI-P: A zero-shot setting where MMSI is directly evaluated on the test set without additional fine-tuning.
    \item MMSI-T: MMSI trained under the same small and big setups as mT5-T for fair comparison.
\end{itemize}

The overall results are summarized in Table~\ref{tab:cmdl}.

It is worth noting that the test set remains unchanged across all experiments to ensure comparability. The results in Table~\ref{tab:cmdl} demonstrate that MMSI outperforms mT5-T under both training and zero-shot conditions, showing strong robustness across datasets and scales. Particularly, on the small dataset, MMSI-T achieves substantial gains in both ImpAcc and ImpErr compared to the zero-shot setting, highlighting the effectiveness of role-aware modeling in complex sentencing scenarios and the suitability of the small dataset for this task. Moreover, when trained on the big dataset, MMSI exhibits reduced performance relative to its zero-shot baseline, further validating the quality and task alignment of our IMLJP dataset.

\section{Joint Multi-Task Model with Classification and Regression}
\label{appsec:E_joint_MMSI}
To unify previously separated learning stages, we propose a joint multi-task learning model that integrates both classification and regression objectives into a single end-to-end architecture. 

The model consists of two independent BERT encoders: one processes the FD, and the other processes the CV$_d$. The FD encoder extracts semantic features from the FD and outputs a classification logit vector. This vector is converted into soft probabilities via softmax and then projected into the hidden space to form a task-aware feature representation. This projected feature is treated as auxiliary semantic knowledge from the classification task. Simultaneously, the CV$_d$ encoder produces token-level hidden states along with multi-head attention maps. A custom attention mechanism is applied to the last-layer hidden states, where the attention weights are modulated by the actual self-attention scores. The resulting refined [CLS] embedding is then fused with the projected classification-derived feature, enabling the model to jointly capture factual and decision-level semantics. The fused representation is passed through a classification head to produce the final prediction and a regression head to estimate the interpretability score.

The model is trained in a multi-task manner using a weighted combination of the classification loss and regression loss:
\begin{equation*}
	\mathcal{L}_{\text{total}} = \alpha \cdot \mathcal{L}_{\text{cls}} + \beta \cdot \mathcal{L}_{\text{reg}},
\end{equation*}
where
\begin{align*}
	\mathcal{L}_{\text{cls}} &= \text{CrossEntropyLoss}(\hat{y}_{\text{cls}}, y_{\text{cls}}), \\
	\mathcal{L}_{\text{reg}} &= \text{MSE}(\hat{y}_{\text{reg}}, y_{\text{reg}}),
\end{align*}
with $\hat{y}_{\text{cls}}$ and $\hat{y}_{\text{reg}}$ denoting the predicted class logits and regression outputs, respectively. The scalars $\alpha, \beta > 0$ control the relative importance of the classification and regression tasks during training.

This joint optimization scheme enables the model to balance performance across both objectives and supports flexible task emphasis by tuning $\alpha$ and $\beta$. Table~\ref{tab:joint-ablation} presents the joint multi-task model's performance under various loss weight combinations $(\alpha, \beta)$. All results are reported as the mean over 10 independent runs to ensure robustness.

\begin{table*}[ht]
	\centering
	\caption{Performance of the joint multi-task model under different loss weight combinations.}
	\label{tab:joint-ablation}
	\begin{tabular}{c|cccc|ccc}
		\hline
		Loss Weights & 
		\multicolumn{4}{c|}{Guilt Inference Task} & 
		\multicolumn{3}{c}{Prison Prediction Task} \\
		\hline
		($\alpha$, $\beta$) & Acc. & P. & R. & F1 & ImpScore & ImpAcc & ImpErr \\
		\hline
		(0.00, 1.00) & 0.4857 & 0.3480 & 0.4857 & 0.3856 &   0.7336 &  0.4555 &  0.0727 \\
		(0.01, 1.00) & 0.5720 & 0.5248 & 0.5720 & 0.5354 & 0.7555 & 0.4638 & 0.0717 \\
		(0.05, 1.00) & 0.6063 & 0.6136 & 0.6063 & 0.5979 & 0.7761 & 0.4862 & 0.0663 \\
		(0.10, 1.00) & 0.5939 & 0.5979 & 0.5939 & 0.5717 & 0.7417 &  0.4379 & 0.0745 \\
		(0.20, 1.00) & 0.5328 & 0.5313 & 0.5328 & 0.5028 & 0.7132 &  0.4235 & 0.0774 \\
		(0.40, 1.00) & 0.5745 & 0.5754 & 0.5745 & 0.5463 & 0.7145 &  0.4091 & 0.0804 \\
		(0.60, 1.00) & 0.6071 & 0.6142 & 0.6071 & 0.5959 & 0.7470 &  0.4693 & 0.0721 \\
		(0.80, 1.00) & 0.6076 & 0.5887 & 0.6076 & 0.5824 & 0.7551 &  0.4650 & 0.0698 \\
		\hline
		(1.00, 1.00) & 0.5550 & 0.5257 & 0.5550 & 0.5205 & 0.7211 & 0.4209 & 0.0768 \\
		\hline
		(1.00, 0.80) & 0.5948 & 0.5746 & 0.5948 & 0.5752 & 0.7684 &  0.4770 & 0.0691 \\
		(1.00, 0.60) & 0.5973 & 0.5857 & 0.5973 & 0.5538 & 0.7216 &  0.4256 & 0.0734 \\
		(1.00, 0.40) & 0.6723 & 0.6840 & 0.6723 & 0.6670 & 0.7718 &  0.4818 & 0.0697 \\
		(1.00, 0.20) & 0.6762 & 0.6305 & 0.6762 & 0.6411 & 0.7173 &  0.4328 & 0.0780 \\
		(1.00, 0.10) & 0.8197 & 0.8232 & 0.8197 & 0.8192 & \textbf{0.7826} & \textbf{0.4979} & 0.0658 \\
		(1.00, 0.05) & 0.7833 & 0.7889 & 0.7833 & 0.7817 & 0.7664 & 0.4724 & \textbf{0.0653} \\
		(1.00, 0.01) & 0.8713 & 0.8750 & 0.8713 & 0.8707 & 0.7638 &  0.4780 & 0.0645 \\
		(1.00, 0.00) & \textbf{0.8987} & \textbf{0.8992} & \textbf{0.8987} & \textbf{0.8987} & 0.0000 &  0.0000 & 0.1887 \\
		
		\hline
	\end{tabular}
\end{table*}

When $\alpha=0$ (only regression), the guilt inference task accuracy is low, reflecting poor ability to distinguish guilt roles. However, the prison prediction metrics remain moderate, indicating reasonable regression performance. Increasing $\alpha$ while fixing $\beta=1.0$ improves classification metrics significantly (e.g., at $(\alpha, \beta) = (0.05, 1.00)$, F1=0.5979), while the prison prediction metrics also improve slightly. This demonstrates that incorporating classification loss helps the model better identify offender roles without sacrificing prison prediction quality. When $\beta=0$ (only classification), the model achieves very high guilt inference accuracy but fails to produce meaningful prison predictions. The best overall balance appears when both tasks are jointly trained with moderate weights (e.g., $\alpha=1.00$, $\beta$ between 0.05 and 0.10), yielding high accuracy in offender role classification and strong prison prediction metrics simultaneously.

These results indicate that the joint multi-task learning framework effectively leverages shared semantic representations to distinguish primary and secondary offenders (classification) and accurately estimate prison terms (regression), with loss weights $\alpha$ and $\beta$ serving as key hyperparameters to balance these objectives.

Among all evaluated settings, the configuration with $(\alpha=1.00, \beta=0.10)$ yields the most favorable trade-off between the two tasks. Under this setting, the model achieves high classification performance for guilt inference and demonstrates superior predictive accuracy in the prison term regression task. We therefore select this setting as the representative result for joint training and compare it with other approaches in the main text. These results suggest that the model is capable of reliably distinguishing between principal and accomplice roles while simultaneously producing precise sentencing predictions, indicating strong practical reliability.

\section{Additional Experiments Supporting Main Results}
\label{appsec:F_additional_experiments}
% To further validate the effectiveness of various input contents and the proposed masking method, tests were conducted on the baselines using different input data (FDs and CV$_d$s). Additionally, the baselines underwent enhanced testing with a label broadcast embedding model (F). The comparative experimental results presented in Table \ref{tab:baseline}. All results are reported as the mean over 10 independent runs to ensure robustness.

To further validate the effectiveness of different input contents and the proposed masking method, we evaluate the baselines using two input settings (FDs and CV$_d$s). Additionally, the baselines are further tested with a label broadcast embedding model (F). The comparative experimental results are presented in Table~\ref{tab:Baseline}. All results are reported as the mean over 10 independent runs to ensure robustness.

For the SOTA LLMs (Claude-3, GPT-3.5, GPT-4o, LLaMA3-70B, Gemini-1.5, and DeepSeek-V3), predictions are obtained via public APIs using standardized prompt templates. The detailed prompt construction is described in Appendix~\ref{sec:K_prompt_info}. Among the baseline models reported in the main text, LADAN and NeurJudge are evaluated following the original input settings described in their respective publications, namely using FD-only inputs. For other models such as Legal-BERT and Lawformer, we report results based on the better-performing input (FDs or CV$_d$s) identified during validation. The SOTA LLMs achieve their best performance when both FD and CV$_d$ are provided jointly as input, consistent with the prompt design.

\begin{table*}[!ht]
	\centering
	\caption{Comparison among the prison prediction results produced by the existing and proposed methods using various data sources and innovative techniques.}
	\label{tab:baseline}
	\begin{tabular}{lcccccc}
		\toprule
		\multirow{2}{*}{Method} & \multicolumn{3}{c}{Unmasked Data} & \multicolumn{3}{c}{Masked Data} \\
		\cmidrule(lr){2-4} \cmidrule(lr){5-7}
		& ImpScore & ImpAcc & ImpErr & ImpScore & ImpAcc & ImpErr \\
		\midrule
		LADAN+FD & 0.2878 & 0.1482 & 0.1519 & 0.3231 & 0.1721 & 0.1481 \\
		LADAN+CV$_d$ & 0.3086 & 0.1550 & 0.1499 & 0.3280 & 0.1608 & 0.1480 \\
		NeurJudge+FD & 0.4327 & 0.2432 & 0.1398 & 0.4312 & 0.2409 & 0.1418\\
		NeurJudge+CV$_d$ & 0.4200 & 0.2401 & 0.1412 & 0.4200 & 0.2401 & 0.1412\\
		\midrule
		NeurJudge+FDF & 0.4739 & 0.2682 & 0.1288 & 0.4752 & 0.2675 & 0.1282\\
		NeurJudge+CV$_d$F & 0.4728 & 0.2560 & 0.1258 & 0.4678 & 0.2556 & 0.1275\\	
		\midrule
		Legal BERT+FD & 0.4383 & 0.2126 & 0.1307 & 0.4398 & 0.2163 & 0.1304\\
		Legal BERT+CV$_d$ & 0.4347 & 0.2143 & 0.1325 & 0.4404 & 0.2186 & 0.1310\\
		
		Lawformer+FD & 0.4816 & 0.2694 & 0.1236 & 0.6049 & 0.3763 & 0.1076\\
		Lawformer+CV$_d$ & 0.4247 & 0.2651 & 0.1312 & 0.6277 & 0.3780 & 0.1044\\        
		\midrule
		HRN & 0.4884 & 0.1961 & 0.1245 & 0.4855 & 0.2812 & 0.1230 \\
		MMSI & 0.6762 & 0.3590 & 0.0863 & 0.7526 & 0.4653 & 0.0680\\
		
		\midrule
        Claude-3+FD& 0.2780& 0.1158 & 0.2513 & - & - & - \\
		Claude-3+CV$_d$ & 0.5194 & 0.2772  & 0.1285 & - & - & - \\
		GPT-3.5+FD & 0.5650 & 0.2814 & 0.1314 & - & - & - \\
		GPT-3.5+CV$_d$ & 0.4773 & 0.2246 & 0.1257 & - & - & - \\
        GPT-4o+FD & 0.5403& 0.2726 & 0.1606 & - & - & - \\
		GPT-4o+CV$_d$& 0.6242 & 0.3553 & 0.1195 & - & - & - \\
        Gemini-1.5+FD & 0.5979 & 0.3250 &0.1047 & - & - & - \\
		Gemini-1.5+CV$_d$ & 0.5075  & 0.2652 & 0.1137 & - & - & - \\    
		Lamma3-70B+FD &0.5828 & 0.2570 & 0.1043 & - & - & - \\
		Lamma3-70B+CV$_d$ &0.5159 & 0.2585 & 0.1158 & - & - & - \\
		DeepSeek-V3+FD & 0.6866 & 0.3751 & 0.0792 & - & - & - \\
		DeepSeek-V3+CV$_d$ & 0.7249 & 0.4266 & 0.0763 & - & - & - \\
		\midrule
        Claude-3+FD+CV$_d$ & 0.3570 & 0.1693 & 0.3721 & - & - & - \\
		GPT-3.5+FD+CV$_d$ & 0.5527 & 0.2719 & 0.1223 & - & - & - \\
        Gemini-1.5+FD+CV$_d$ &0.5731 & 0.2977 & 0.0976 & - & - & - \\
		Lamma3-70B+FD+CV$_d$ &0.5798 & 0.2898 & 0.1085 & - & - & - \\
        GPT-4o+FD+CV$_d$ & 0.6271 & 0.3574 & 0.1185 & - & - & - \\
		DeepSeek-V3+FD+CV$_d$ & 0.7543 & 0.4505 & 0.0655 & - & - & - \\
		\bottomrule
	\end{tabular}
\end{table*}

The results presented in Table~\ref{tab:baseline} further confirm the effectiveness of the proposed MMSI model and provide a detailed comparison with a range of existing methods under different input configurations and enhancement strategies. Across the traditional baselines, LADAN and NeurJudge are evaluated using both FD and CV$_d$. Although CV$_d$ contains more legal reasoning elements, its impact on model performance is mixed: LADAN slightly benefits from CV$_d$ in ImpScore, while NeurJudge remains relatively stable across input types. Legal-BERT and Lawformer benefit more from FD inputs because FDs exhibit greater role-specific variability and sparser expressions than the relatively uniform wording in CV$_d$.

In addition to input variation, injecting label broadcast embeddings notably improved NeurJudge's predictive capabilities. This enhancement led to consistent improvements in both accuracy and error reduction under both FD and CV$_d$ conditions, indicating that explicit guilt-role information, even when indirectly constructed, helps in capturing sentencing patterns more effectively.

The proposed masking strategy, designed to isolate defendant-specific information in multi-defendant scenarios, yields significant benefits for several models. Lawformer, in particular, achieves large performance gains when masking is applied, highlighting the advantage of defendant-focused context modeling. Similarly, HRN demonstrates improvement in accuracy despite its generative nature, suggesting that even for autoregressive models, structural disambiguation through masking contributes positively.

When compared with SOTA LLMs accessed via APIs—namely Claude-3, GPT-3.5, GPT-4o, LLaMA3-70B, Gemini-1.5, and DeepSeek-V3—the MMSI model demonstrates competitive or superior performance. DeepSeek-V3 achieves the best performance among the LLMs when using combined FD and CV$_d$ inputs. Nonetheless, under masked settings, MMSI achieves similar performance, indicating that domain-specific architectural designs and targeted training strategies can match the performance of general-purpose models on structured legal tasks. Furthermore, MMSI relies solely on principal/accomplice labels inferred from FD, using less input information than the SOTA LLMs, which highlights its extensibility. The framework also exhibits strong generalizability, and in several backbone configurations, it outperforms SOTA LLMs, as demonstrated in the experimental results presented in the main text.

Overall, the MMSI outperformed baselines. Its ability to incorporate guilt-role inference, handle multi-defendant contexts, and leverage both factual and reasoning inputs renders it particularly effective for sentence prediction in complex legal scenarios. These results not only validate the design of MMSI but also demonstrate its superiority over both fine-tuned baselines and powerful black-box LLMs.

\section{Attention-Based Case Study}
\label{appsec:G_attention_case_study}

\begin{figure*}[!h]
	\centering
	\includegraphics[width=0.9\linewidth]{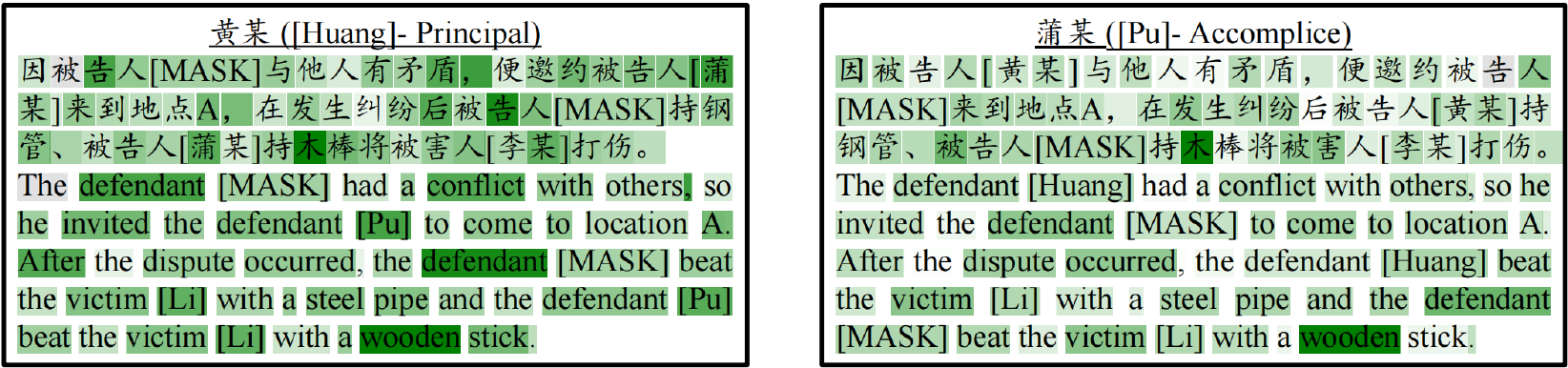}
	\caption{\label{fig:case_study}Attention visualization produced for the guilt inference model based on FDs, which compares defendants with different guilt roles in the same case. The normalized attention weights yielded by the final model layer for each token are shown as a gradient from white to green, with darker shades indicating greater levels of attention. The masked defendant names and their respective guilt roles are displayed in the first row. The left panel represents the principal defendant, whereas the right panel corresponds to the accomplice.}
\end{figure*}

Besides the IG attribution values discussed in the main text, we further enhance interpretability by generating an attention visualization from the model's final layer. As illustrated in Fig.~\ref{fig:case_study}, a typical case is selected to demonstrate the model's attention when distinguishing the responsibilities of principal and accomplice defendants. The inference required in such cases demands rigorous judicial logic and the integration of factors such as the causes and attributions of fatal injuries. Analyzing the facts of crimes poses significant challenges for machine comprehension.

\begin{figure*}[!h]
	\centering
	\fbox{\includegraphics[width=0.9\linewidth]{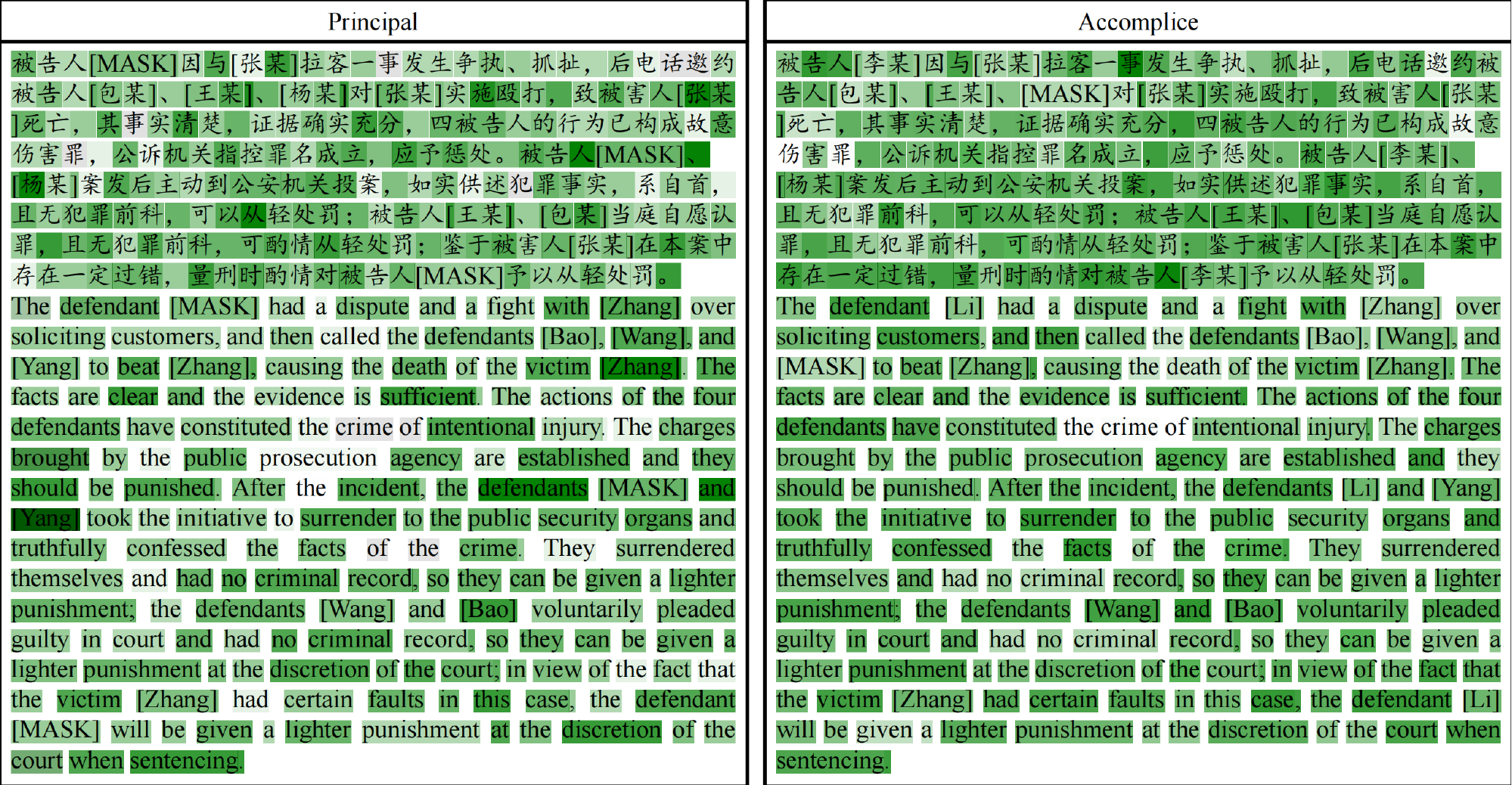}}
	\caption{\label{fig:prision_study}Attention visualization produced for the prison prediction model based on CV$_d$s, which compares defendants with different guilt roles in the same case. The normalized attention weights yielded by the final model layer for each token are shown as a gradient from white to green, with darker shades indicating greater levels of attention. The left panel represents the principal defendant, whereas the right panel corresponds to the accomplice.}
\end{figure*}

In the classification model, the incorporation of the oriented masking mechanism for the target defendant allows the model to focus more on the contextual information surrounding that defendant. Additionally, it enables the model to analyze factors such as the causes of a crime and the tools used in its commission. This further enhances the interpretability of the proposed solution.

Additionally, to assess the effectiveness of the model in the prison prediction task, a single-case attention check was conducted on the regression model (Fig. \ref{fig:prision_study}). The results reveal that for regression tasks, the model effectively incorporates all contextual information contained within the input CV$_d$, including event causes such as disputes and outcomes such as fatalities. Moreover, it demonstrates the ability to identify passages containing sentencing-relevant elements, such as voluntary surrender, prior criminal records, and victim culpability. The attention of the model tends to focus more on the context surrounding masked mentions of the defendant's name. For principals, whose descriptions are more detailed and widely distributed throughout the text, the attention patterns exhibit noticeable clusters. In contrast, the attention paid to accomplices is more evenly distributed, with a smaller variance. This attention behavior highlights one of the internal mechanisms contained in the model for differentiating between the sentencing predictions produced for principals and accomplices.

\section{Attribution-Based Insight into Sentencing Decisions}
\label{appsec:H_word_cloud}
\begin{figure}[!h]
	\centering
	\includegraphics[width=0.8\linewidth]{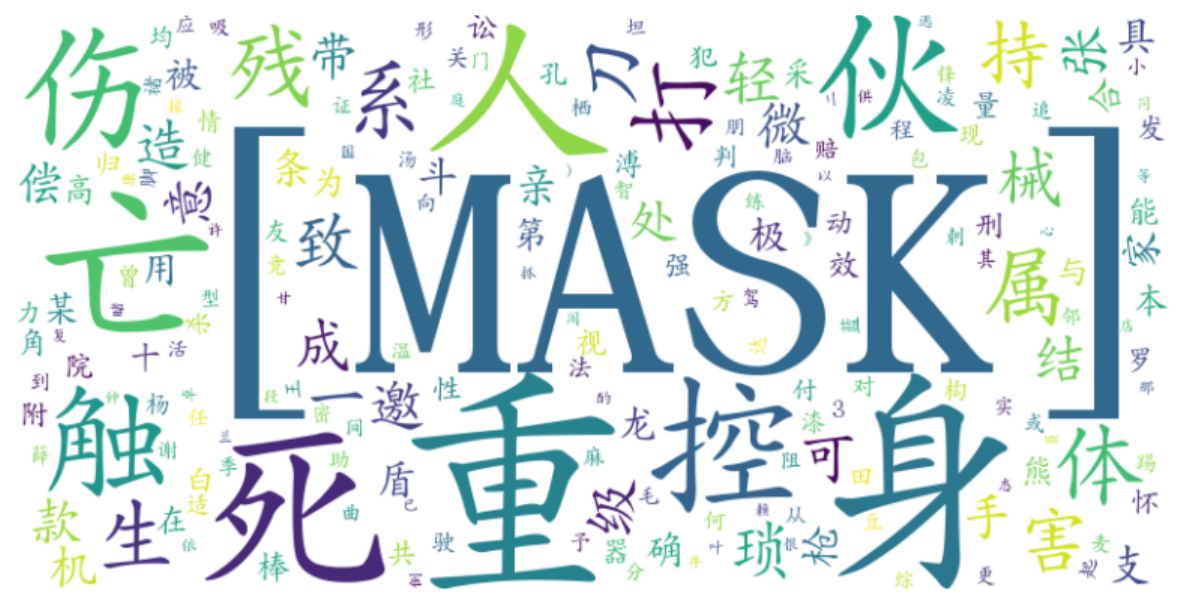}
	\caption{\label{fig:wordcloud}Word cloud of important tokens identified via Integrated Gradients in the prison prediction model. The top-5 tokens with the highest attribution scores were collected from each sample to construct a global view of model attention. Tokens appearing more frequently suggest stronger and more consistent influence on the predicted outputs.
	}
\end{figure}

To further investigate the interpretability of our sentencing regression model, we employ the Integrated Gradients (IG) method to analyze the contribution of individual input tokens. However, due to the inherently continuous nature of the regression task, the interpretability of single-token attributions is often not immediately intuitive. To address this, we identify the top-5 most influential tokens based on their IG scores and visualize them using word clouds in Fig. \ref{fig:wordcloud}.

\begin{table*}[h]
	\centering
	\caption{Comparison among the results of ablation experiments.}
	\label{tab:ablation}
	\begin{tabular}{l|ccc}
		\hline
		Models & ImpScore ($\pm$95\%CI) & ImpAcc ($\pm$95\%CI) & ImpErr ($\pm$95\%CI)\\
		\hline
		MMSI without M & 0.6763 ($\pm$0.0261)& 0.3590 ($\pm$0.0268)& 0.0862 ($\pm$0.0038)\\
		% \hline
		MMSI without F & 0.6722 ($\pm$0.0206) & 0.3718 ($\pm$0.0245)& 0.0879 ($\pm$0.0025)\\
		% \hline
		MMSI & 0.7526 ($\pm$0.0119)& 0.4653 ($\pm$0.0160)& 0.0680 ($\pm$0.0027)\\
		\hline		
        \multicolumn{4}{r}{\footnotesize{$\pm$95\% CI: 95\% confidence interval}}\\
	\end{tabular}
\end{table*}

The visualization reveals that the model tends to assign high importance to tokens related to severe case outcomes, such as \textquotedblleft serious injury$\textquotedblright$ and \textquotedblleft death\textquotedblright, as well as to expressions of collective criminal behavior, such as \textquotedblleft assault\textquotedblright, and references to weapons such as \textquotedblleft knife\textquotedblright. These observations suggest that the model effectively captures key factual indicators relevant to sentencing decisions.

Notably, we observe that our proposed use of [MASK] tokens significantly affects the model's predictions in prison prediction tasks. In particular, the model demonstrates sensitivity to the positions in which [MASK] tokens appear, which highlights their utility in emphasizing the behavioral roles of individual defendants within the factual narrative.

\section{Ablation Experiments}
\label{appsec:I_ablation}
To further assess the impacts of the various innovations contained within the complete solution, ablation experiments were conducted; specifically, the oriented masking method (M) and the model enhancement ability of label broadcasting (F) were examined. The experimental results are presented in Table \ref{tab:ablation}.

The results indicate that the proposed model exhibits strong prison prediction performance, benefiting from both the masking mechanism and the enhanced model embedding acquired through label broadcasting. Additionally, the effectiveness of the model in the upstream task relies on the presence of the masking mechanism, underscoring the importance of both the masking approach and the multistage information transmission scheme.

\section{Sensitivity Analysis}
\label{appsec:J_sensitivity}
To conduct a sensitivity analysis, the performance of the model is evaluated across various hyperparameter configurations, testing its robustness and sensitivity to parameter changes. Specifically, the following parameters are varied:

1. \textbf{Epochs}: The number of training epochs is varied within \([5, 10, 20, 50]\) to determine how an increased number of iterations influences the accuracy and convergence of the model. This approach helps identify the optimal number of epochs, thereby balancing performance plateauing and overfitting.

2. \textbf{Learning Rate}: Learning rates within the range \([0.00001, 0.0001, 0.001, 0.01, 0.1]\) are adjusted to evaluate the sensitivity of the model to this parameter, which controls the weight update rate. Higher learning rates allow us to observe convergence speed, whereas lower rates highlight stability and accuracy improvements.

3. \textbf{Dropout Rate}: Dropout rates of \([0.1, 0.3, 0.5]\) are explored to assess the effects of regularization, thereby balancing overfitting prevention with potential training accuracy reductions.

\begin{figure}[!ht]
	\centering
	\includegraphics[width=0.95\linewidth]{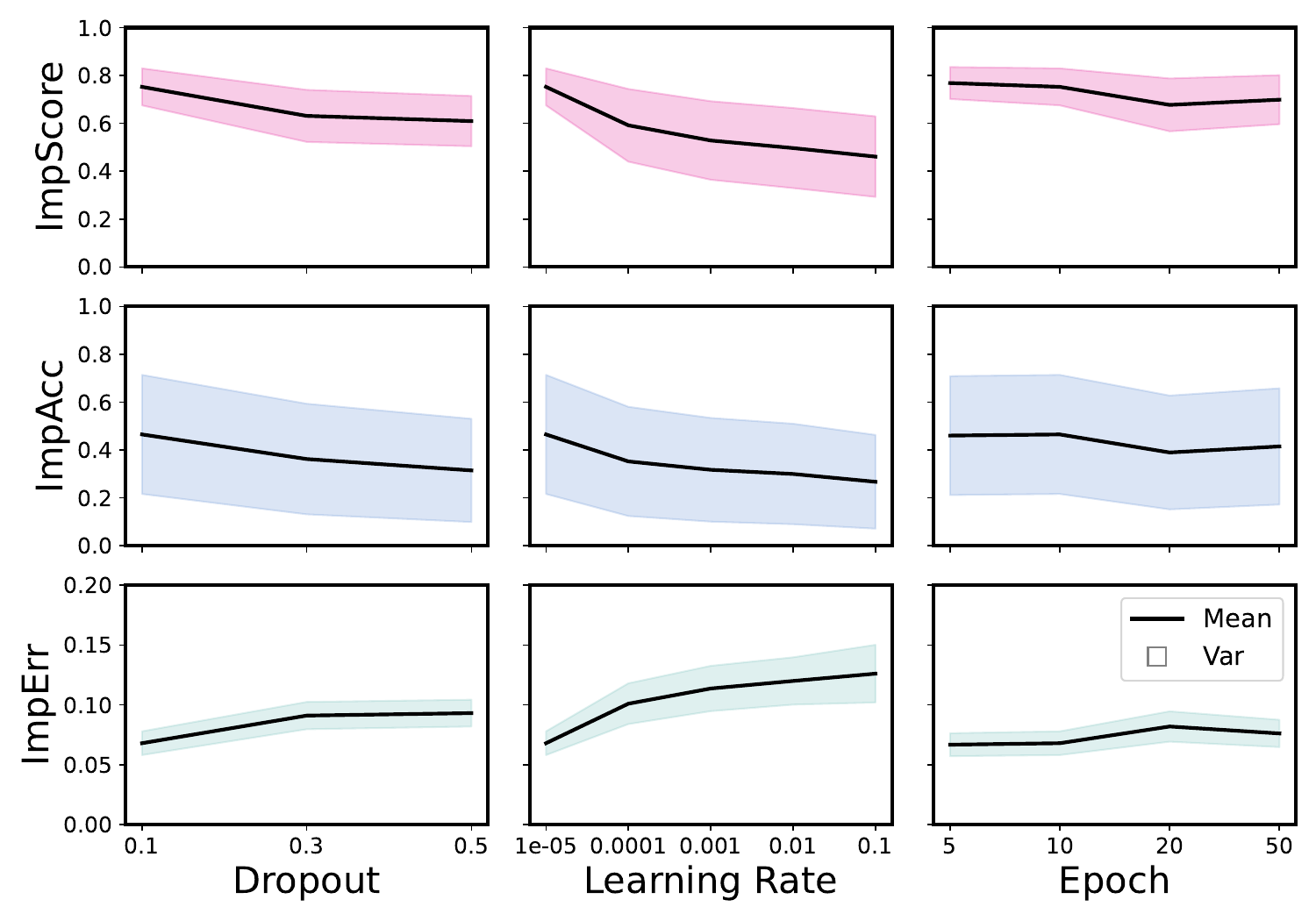}
	\caption{\label{fig:sensitive_analysis}The analysis used the baseline parameters, i.e., \( \text{epochs} = 10 \), \( \text{learning rate} = 2 \times 10^{-5} \), \( \text{batch size} = 16 \), and \( \text{dropout rate} = 0.1 \), with parameter adjustments evaluated across three performance metrics.}
\end{figure}

Owing to computational resource constraints, input batch size testing is not conducted; however, tests involving batch sizes of 16 and 32 show no significant performance differences. The effects of parameter adjustments on the resulting accuracy, stability, and convergence are presented in Fig.~\ref{fig:sensitive_analysis}, where the baseline parameters are used for comparison purposes.

The experimental results indicate that increasing the dropout rate led to a decrease in model accuracy. Additionally, the model yielded stable, highly precise predictions after a relatively small number of training epochs. The model is sensitive to the learning rate, with larger values causing instability and accuracy degradation. In summary, the model demonstrates strong robustness.

\section{API and Prompt-based Training Model Prompt Information}
\label{sec:K_prompt_info}

To ensure a fair comparison with LLM baselines accessed via APIs (e.g., Claude-3, GPT-3.5, GPT-4o, LLaMA3-70B, Gemini-1.5 and DeepSeek-V3) as well as prompt-based models trained via fine-tuning (e.g., Finetune-CoT, mT5), we adopt a standardized prompt format for prison prediction. Notably, for the HRN model, whose code is publicly available, we use its original prompt without any modification. This format directs the model to infer the sentencing outcome based on the FD and CV$_d$ corresponding to a specific defendant.

We consider three prompting scenarios depending on the available input information:

\begin{table*}[ht]
	\centering
	\caption{Performance comparison of prompting strategies under the MMSI framework.}
	\label{tab:prompt_enhance}
	\begin{tabular}{l|ccc}
		\hline
		\textbf{Prompt Strategy} & \textbf{ImpScore ($\pm$95\%CI)} & \textbf{ImpAcc ($\pm$95\%CI)} & \textbf{ImpErr ($\pm$95\%CI)}\\
		\hline
		Single-stage& 0.7543 ($\pm$0.0052) & 0.4505 ($\pm$0.0097) & 0.0655 ($\pm$0.0021) \\
		Two-stage& 0.7782 ($\pm$0.0052) & 0.5135 ($\pm$0.0098) & 0.0542 ($\pm$0.0015) \\
		\hline
		\multicolumn{4}{r}{\footnotesize{$\pm$95\% CI: 95\% confidence interval}}\\
	\end{tabular}
\end{table*}
%\begin{CJK}{UTF8}{gbsn} 
(1) Prompt based on FD and CV$_d$:
\begin{mdframed}
\small
Given the following fact description and court view of the case, infer the specific term of imprisonment (in months) for the defendant \{name\}.  

Fact description: \{fd\}.  

Court view: \{cvd\}.  

Please provide the exact number of months in the prison term for the defendant \{name\}.
\end{mdframed}

(2) Prompt based on FD only:
\begin{mdframed}
\small
Given the following fact description of the case, infer the specific term of imprisonment (in months) for the defendant \{name\}. 

Fact description: \{fd\}. 

Please provide the exact number of months in the prison term for the defendant \{name\}.
\end{mdframed}

(3) Prompt based on CV$_d$ only:
\begin{mdframed}
\small
Given the following court view of the case, infer the specific term of imprisonment (in months) for the defendant \{name\}. 

Court view: \{cvd\}. 

Please provide the exact number of months in the prison term for the defendant \{name\}.
\end{mdframed}

In these prompts:
\begin{itemize}
    \item \texttt{\{name\}} represents the name sequence of the target defendant;
    \item \texttt{\{fd\}} corresponds to the FD sequence;
    \item \texttt{\{cvd\}} corresponds to the CV$_d$ sequence.
\end{itemize}

The model is expected to return a numerical value indicating the predicted prison term in months. All prompts are applied uniformly across prompt-based and API-based models to ensure consistency and fairness in experimental comparison.

\section{Enhancing LLM Capabilities under the MMSI Framework}
\label{appsec:L_multi-stage_prompt}
To further enhance reasoning ability in sentencing prediction, this work explores multi-stage prompting with LLMs under the proposed MMSI framework. Specifically, MMSI is grounded in both FD and CV$_d$: the model first performs FD-to-guilt inference to determine role-specific liabilities, and subsequently integrates these results into CV$_d$-to-prison prediction. Experimental results demonstrate that DeepSeek-V3, an industry-level Chinese LLM, achieves the best performance under this FD–CV$_d$ integration, motivating the design of structured multi-stage reasoning strategies that better align with the legal decision-making process.

We consider two prompting strategies: 
\begin{itemize}
	\item \textbf{Single-stage Prompt}: Directly maps facts and court statements into sentencing prediction, which is consistent with Section~\ref{sec:K_prompt_info} (Prompt based on FD and CV$_d$).
	\item \textbf{Two-stage Prompt}: First predicts the defendant's role (principal vs. accomplice), and then predicts sentencing based on this inferred role. 
\begin{mdframed}
\small
\# Round 1: Role inference

User: Given the following fact description, infer whether the defendant \{name\} is a principal or an accomplice.

Fact description: \{fd\}.

\medskip
\hrule
\medskip

\# Round 2: Prison prediction

User: Based on the above principal–accomplice information and the following court view, infer the specific term of imprisonment (in months) for the defendant \{name\}. 

Court view: \{cvd\}.

Please provide the exact number of months in the prison term for the defendant \{name\}.
\end{mdframed}

\end{itemize}

In these prompts:
\begin{itemize}
	\item \texttt{\{name\}} represents the name sequence of the target defendant;
	\item \texttt{\{fd\}} corresponds to the FD sequence;
	\item \texttt{\{cvd\}} corresponds to the CV$_d$ sequence.
\end{itemize}

We evaluate these three prompting strategies on DeepSeek-V3, and the corresponding results are presented in Table \ref{tab:prompt_enhance}.

Results demonstrate that transferring our specially designed two-stage MMSI framework to LLMs through multi-stage prompting significantly improves both accuracy and interpretability.

\ifCLASSOPTIONcaptionsoff
  \newpage
\fi

\bibliography{reference}

\end{appendices}

\end{document}